%% file: main_tech_report.tex
\documentclass[10pt, logo, copyright]{nvidiatechreport}

\usepackage{mdframed}
\usepackage[utf8]{inputenc} 
\usepackage[T1]{fontenc}    

\usepackage{amsfonts}       
\usepackage{nicefrac}       
\usepackage{microtype}      
\usepackage[dvipsnames]{xcolor}         
\usepackage{multirow}
\usepackage{multicol}
\usepackage{graphicx}
\usepackage[numbers]{natbib}
\usepackage{tabto}
\usepackage{xspace}
\usepackage{amsmath}
\usepackage{adjustbox}
\usepackage{enumitem}
\usepackage{wrapfig}
\usepackage{dblfloatfix}
\usepackage{caption}
\usepackage{footmisc}

\usepackage{float}

\usepackage{natbib}
\input{math_commands.tex}

\usepackage{amsmath}
\usepackage{amssymb}
\usepackage{mathtools}
\usepackage{amsthm}
\usepackage{enumitem}
\usepackage{multirow}
\usepackage{booktabs}

\usepackage[capitalize,noabbrev]{cleveref}

\usepackage{algorithm}
\usepackage{algorithmicx}
\usepackage{algpseudocode}
\usepackage{xspace}
\usepackage{csquotes}
\usepackage{listings}
\usepackage{xcolor}

\usepackage{multirow}
\usepackage{multicol}
\usepackage{enumitem}
\usepackage[most,skins,theorems]{tcolorbox}
\tcbset{
  aibox/.style={
    width=\linewidth,
    top=8pt,
    bottom=4pt,
    colback=blue!6!white,
    colframe=black,
    colbacktitle=black,
    enhanced,
    center,
    attach boxed title to top left={yshift=-0.1in,xshift=0.15in},
    boxed title style={boxrule=0pt,colframe=white,},
  }
}

\usepackage{xcolor}
\usepackage{hyperref}

\definecolor{nvidiagreen}{HTML}{76B900}

\hypersetup{
    colorlinks=true,
    urlcolor=nvidiagreen,
    linkcolor=nvidiagreen,
    citecolor=nvidiagreen
}

\newtcolorbox{AIbox}[2][]{aibox,title=#2,#1}

\definecolor{lightgray}{gray}{0.95} 
\definecolor{darkblue}{rgb}{0,0,0.6} 
\definecolor{nvgreen}{cmyk}{50, 0, 100, 0}

\newcommand{\MethodName}{ToolOrchestra}
\newcommand{\DataName}{ToolScale}

\title{{\MethodName}: Elevating Intelligence via Efficient Model and Tool Orchestration}

\author{%
  Hongjin Su$^{*}$$^{1,2}$
  Shizhe Diao$^{*}$$^{1}$ \quad
  Ximing Lu$^{1}$ \quad
  Mingjie Liu$^{1}$ \quad
  Jiacheng Xu$^{1}$ \quad
  Xin Dong$^{1}$ \quad
  \textbf{Yonggan Fu}$^{1}$ \quad 
  \textbf{Peter Belcak}$^{1}$ \quad
  \textbf{Hanrong Ye}$^{1}$ \quad
  \textbf{Hongxu Yin}$^{1}$ \quad
  \textbf{Yi Dong}$^{1}$ \quad
  \textbf{Evelina Bakhturina}$^{1}$ \quad
  \textbf{Tao Yu$^{2}$} \quad
  \textbf{Yejin Choi}$^{1}$ \quad
  \textbf{Jan Kautz}$^{1}$ \quad
  \textbf{Pavlo Molchanov}$^{1}$\\
$^1$NVIDIA, $^2$University of Hong Kong
}
\newcommand{\Tau}{\mathrm{T}}

\correspondingauthor{X}

\begin{abstract}
\textbf{Abstract:} Large language models are powerful generalists, yet solving deep and complex problems such as those of the Humanity’s Last Exam (HLE) remains both conceptually challenging and computationally expensive.
We show that small orchestrators managing other models and a variety of tools can both push the upper bound of intelligence and improve efficiency in solving difficult agentic tasks.
We introduce {\MethodName}, a method for training small orchestrators that coordinate intelligent tools. 
{\MethodName} explicitly uses reinforcement learning with outcome-, efficiency-, and user-preference-aware rewards.
Using {\MethodName}, we produce Orchestrator, an 8B model that achieves higher accuracy at lower cost than previous tool-use agents while aligning with user preferences on which tools are to be used for a given query.
On HLE, Orchestrator achieves a score of 37.1\%, outperforming GPT-5 (35.1\%) while being 2.5x more efficient.
On $\tau^2$-Bench and FRAMES, Orchestrator surpasses GPT-5 by a wide margin while using only about 30\% of the cost.
Extensive analysis shows that Orchestrator achieves the best trade-off between performance and cost under multiple metrics, and generalizes robustly to unseen tools. 
These results demonstrate that composing diverse tools with a lightweight orchestration model is both more efficient and more effective than existing methods, paving the way for practical and scalable tool-augmented reasoning systems.
\vspace{2mm}
\newline
\href{https://github.com/NVlabs/ToolOrchestra/}{\textcolor{nvidiagreen}{Code}} \quad\quad
\href{https://huggingface.co/nvidia/Orchestrator-8B}{\textcolor{nvidiagreen}{Model}} \quad\quad
\href{https://huggingface.co/datasets/nvidia/ToolScale}{\textcolor{nvidiagreen}{Data}} \quad\quad
\href{https://research.nvidia.com/labs/lpr/ToolOrchestra}{\textcolor{nvidiagreen}{Webpage}}

\end{abstract}

\begin{document}
\maketitle


\begin{figure*}[h]
\centering

\begin{minipage}[t]{0.67\textwidth}
    \centering
    \includegraphics[width=\linewidth]{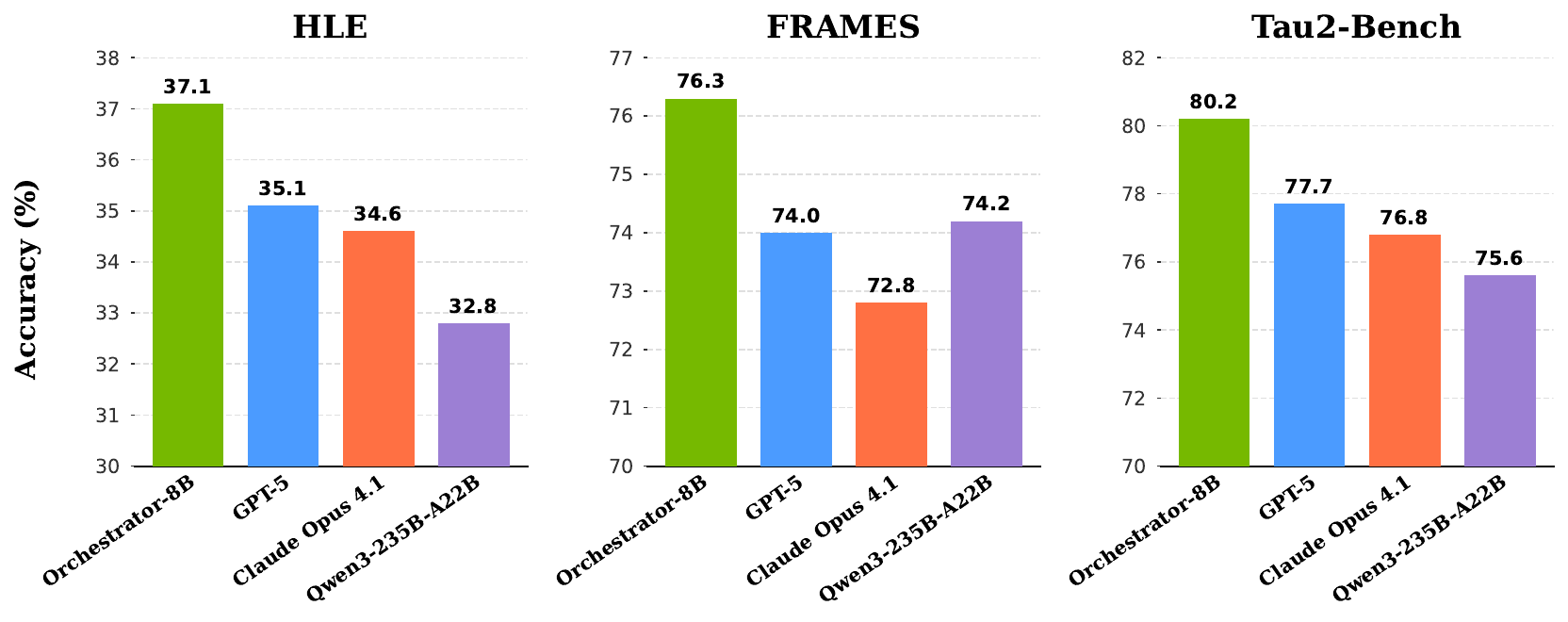}
\end{minipage}
\hfill
\begin{minipage}[t]{0.32\textwidth}
    \centering
    \includegraphics[width=\linewidth]{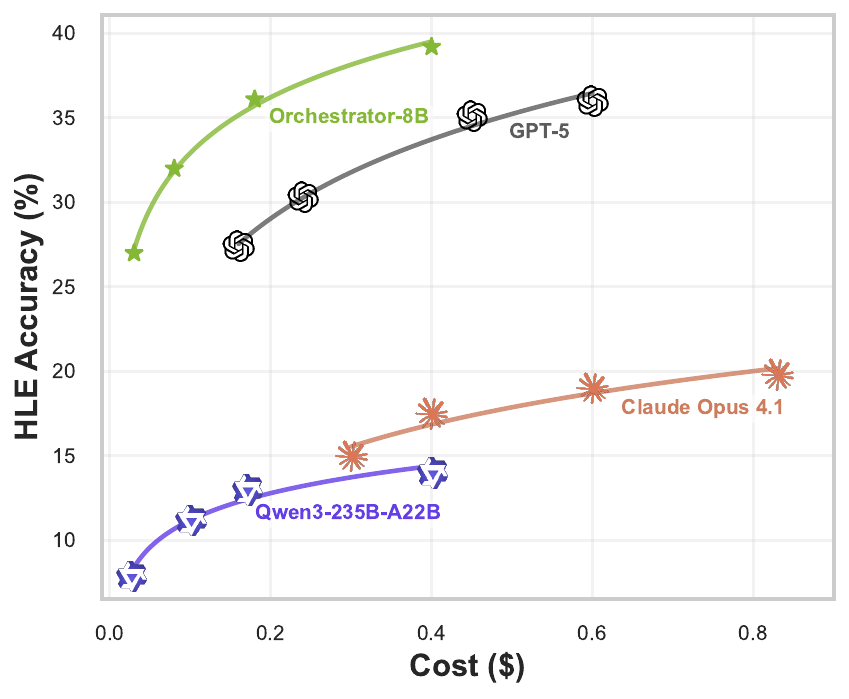}
\end{minipage}

\vspace{-6pt}
\caption{{\MethodName} shows consistently strong performance on HLE, FRAMES, and $\tau^2$-Bench with superior cost efficiency.
}
\label{fig:combined}
\vspace{-10pt}
\end{figure*}

\section{Introduction}
Large language models (LLMs) have been reported to have made remarkable strides towards superhuman intelligence but remain of limited utility in complex agentic tasks such as those posed by the Humanity’s Last Exam (HLE)~\citep{phan2025humanity}.
Tool use is a promising avenue for the extension of their capabilities beyond what can be learned from the training data.
By calling on external resources through search engines and code interpreters, tool use has been shown to enhance accuracy and reduce hallucinations~\citep{qintoolllm, schick2023toolformer, qin2024tool, gehring2024rlef, qian2024tell, yu2024steptool, goldie2025synthetic, zhang2025nemotron, qian2025toolrl}.

Prior research on tool-use agents has primarily focused on equipping a single powerful model with utility tools such as web search or calculators. While effective in many scenarios, this approach underutilizes the potential of tools: humans, when reasoning, routinely extend themselves by calling upon resources of greater-than-human intelligence, from domain experts to sophisticated processes and software systems.
Motivated by this observation, we propose the \textit{orchestration paradigm}. Under this paradigm, intelligence emerges not from a monolith but from a composite system.
At the center of the system lies an \textit{orchestrator} model, whose responsibility is to invoke the right tools for the given task, and to do so in the right order to accomplish the task.
The crucial difference to the standard monolithic setup featuring a single powerful model is that in addition to deterministic utilities such as web search functions and code interpreters, models of various capabilities are made available to the orchestrator as \textit{intelligent tools}. 
The use of tools of different levels of intelligence comes at varying costs, and the challenge for the orchestrator is then to dynamically decide on which tools to invoke in order to solve the task while respecting user preferences for various tools and minimizing the cost.
By delegating narrowed-down sub-problems of a larger effort requiring intelligence to intelligent tools instead of handling the entire effort by a single generalist, orchestration teems with the promise of exhibiting higher intelligence than any of the system's tools and leading monolithic solutions alike.

One approach to implementing the orchestrator paradigm is to employ a language model as the orchestrator and allow it to invoke stronger models only when it deems it necessary. 
This can be done naively by \textit{prompting} an off-the-shelf language model or by \textit{training} a general-purpose orchestrator.
For the former, we find that relying on straightforward model prompting is brittle and introduces systemic biases. 
As shown in Figure~\ref{fig:imbalanced-tool-calls} (left and middle),
GPT-5 disproportionately delegates tasks to GPT-5-mini, while Qwen3-8B defers to GPT-5 at a markedly higher rate.
This illustrates two present issues of prompting in the context of complex tool orchestration: (i) the overuse of developmentally-related variants of oneself, i.e.,  \textit{self-enhancement bias}~\citep{zheng2023judging}, and (ii) defaulting to the strongest available tool regardless of the cost or relative utility (see  Appendix~\ref{sec:app:pilot_study} for more details and \S{}\ref{sec:exp_settings} for a thorough comparison to baselines).
As such, we conclude that the scenarios in which an orchestrating model may call on models and tools of capabilities both inferior and superior to its own are idiosyncratic in the context of model tool calling and warrant their own approach to training.
In addition, controllability in tool-use agents remains underexplored along two axes: cost–efficiency and user preferences (cf. \S\ref{sec:related_work}).

\begin{figure}[t]
    \centering
    \includegraphics[width=\linewidth]{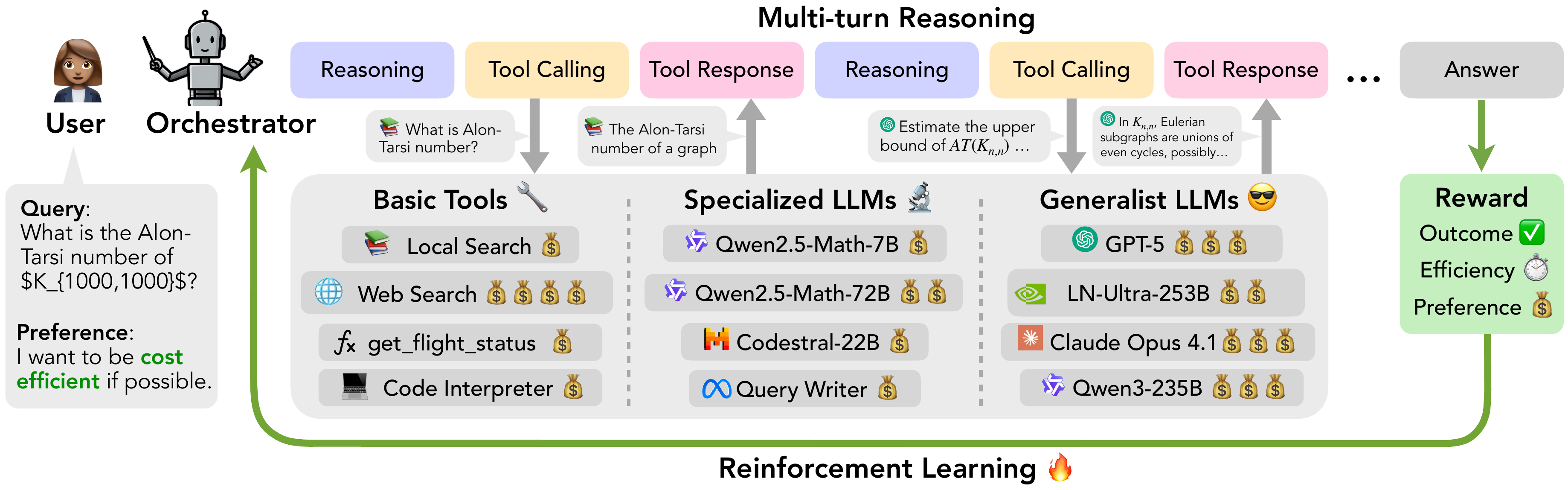}
    \vspace{-20pt}
    \caption{Overview of Orchestrator. Given a task, Orchestrator alternates between reasoning and tool calling in multiple turns to solve it. Orchestrator interacts with a diverse tool set, including basic tools (web search, functions such as \texttt{get\_flight\_status}, etc.), specialized LLMs (coding models, math models, etc.) and generalist LLMs (GPT-5, Claude Opus 4.1, etc.). In training under {\MethodName}, Orchestrator is jointly optimized by outcome, efficiency and preference rewards via reinforcement learning.}
    \label{fig:fig1}
    \vspace{-15pt}
\end{figure}

\begin{wrapfigure}{r}{0.4\columnwidth} 
  \vspace{-8pt}
  \centering
  \includegraphics[width=0.41\columnwidth]{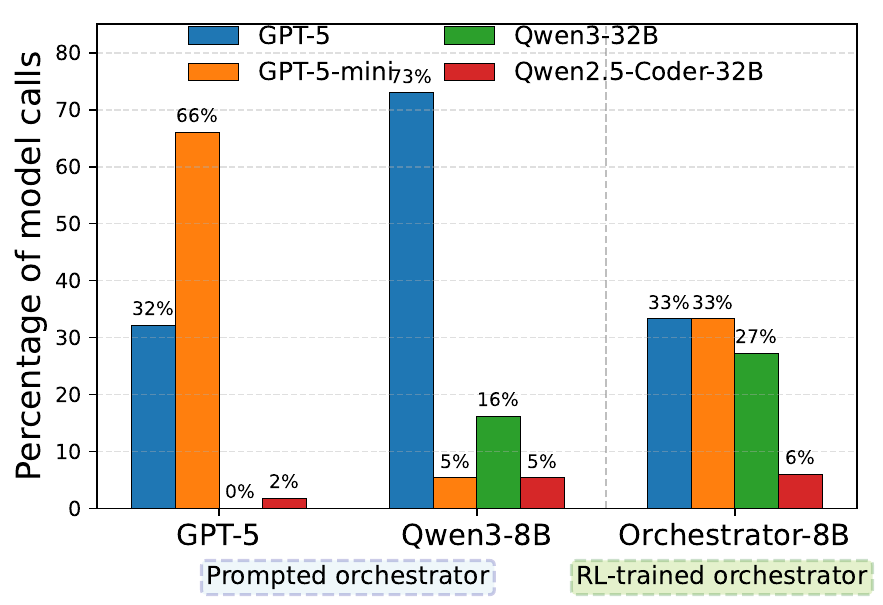}
  \vspace{-12pt}
  \caption{Tool-calling preferences exhibited by a prompted off-the-shelf or RL-trained model.
  GPT-5 tends to call GPT-5-mini most of the time, while Qwen3-8B relies heavily on GPT-5.
  }
  \label{fig:imbalanced-tool-calls}
  \vspace{-12pt}
\end{wrapfigure}

We address these shortcomings by proposing {\MethodName} (shown in Figure~\ref{fig:fig1}), a novel method for training a small language model to act as the orchestrator -- the ``brain'' of a heterogeneous tool-use agent.
Using {\MethodName}, we produce the Orchestrator, an 8B-parameter model trained end-to-end with reinforcement learning (RL) to decide when and how to invoke more intelligent language models and various tools such as web search or code interpreters, and how to combine them in multi-turn reasoning.
Our reward design balances three objectives -- correctness of the final outcome, efficiency in resource usage, and alignment with user preferences --
to yield a cost-effective and user-controllable tool-use policy. 
To aid RL training, we build an automatic data synthesis pipeline that generates thousands of verifiable multi-turn tool-use training examples with complex environments across 10 domains. 
We will make the resulting dataset, {\DataName}, publicly available to facilitate further research on tool-use agent training.

In our experiments, we rigorously evaluate the merits of our approach on three challenging tasks. 
On HLE~\citep{phan2025humanity}, a benchmark consisting of difficult questions across many disciplines, we find that Orchestrator substantially outperforms prior methods with far lower computational cost. 
We also test on $\tau^2$-Bench~\citep{barres2025tau}, a function-calling benchmark, where Orchestrator demonstrates the ability to schedule a variety of tools effectively, calling a large model (GPT-5) in only $\sim$40\% of the steps and utilizing cheaper models or tools for the rest, yet still exceeding the performance of an agent that uses the large model for every step. 
Finally, additional evaluations on the FRAMES~\citep{krishna2024factfetchreasonunified}, a factuality reasoning benchmark, provide further evidence of the versatility and robustness of our approach. 
We observe that even though the training and testing tasks differ markedly, the RL-trained Orchestrator adapts its tool-use policy to new challenges, indicating a high degree of general reasoning ability.

Our contributions can be summarized as follows:
(1) We introduce {\MethodName}, a method for training a small language model to serve as the orchestrator of a diverse toolkit, including classical tools and more intelligent models. This dovetails with recent developments in the field testifying that small language models are often sufficiently powerful and far more economical in agentic systems~\citep{belcak2025small,zhao2025llm}.
(2) We develop a novel reward training design that goes beyond accuracy. The resulting Orchestrator is trained end-to-end to balance task outcome correctness, efficiency in cost and latency, and alignment with user cost and tool preferences.
(3) We demonstrate that Orchestrator trained by {\MethodName} achieves state-of-the-art performance on challenging reasoning benchmarks, surpassing frontier models while using only a fraction of their compute and wall-clock time, and that it generalizes robustly to unseen tasks and tools.

\section{Agentic Problem Formulation}

\label{sec:problem_formulation}
\subsection{Task Formulation}

We investigate multi-turn tool-use agentic tasks and formalize them as a Markov Decision Process (MDP) 
$\mathcal{M}=(\mathcal{U},\mathcal{S},\mathcal{A},\mathcal{O},\mathcal{T},\mathcal{Z},r,\rho,\gamma)$ following conventions similar to prior work~\citep{xi2024agentgym,zhou2024archer,xi2025agentgym}.
We are given an instruction $u\in\mathcal{U}$, user action preferences $p = \left(0 \leq p_{a} \leq 1 \text{ for } a \in \mathcal{A}\right)$, an initial state drawn from 
$\rho(\cdot\,|\,u)$, an initial observation $o_0\in\mathcal{O}$, and the environment state space $\mathcal{S}$.
At step $k$, the Orchestrator chooses an action 
$a_k\in\mathcal{A}$ according to a policy 
$\pi_\theta(a_k\,|\,h_k)$ where $h_k=(u,o_0,a_0,o_1,\dots,a_{k-1},o_k)$ 
is the interaction history.
The environment transitions according to 
$\mathcal{T}(s_{k+1}\,|\,s_k,a_k)$ and emits an observation 
$o_{k+1}\sim\mathcal{Z}(\cdot\,|\,s_{k+1},a_k)$.
The actions $a_i$ come at costs $c_i$ and operational latency $l_i$, and the alignment of each action with user preferences is $p_{a_i}$.
After $N$ interaction steps, Orchestrator has traced the trajectory $\tau=h_{N}$ and the environment provides a reward $r(\tau) \in [0,1]$ based on its correctness.
Our goal is to maximize the correctness reward $r(\tau)$ and the overall user preference alignment $\sum p_{a_i}$ while minimizing the total cost $\sum c_i$ and the aggregate latency $\sum l_i$. 

\subsection{Multi-Turn Rollout}

Given a user task, Orchestrator produces a solution via an iterative rollout that interleaves tool use with environment feedback to form a trajectory of turns. 
The rollout is initialized with a predefined system prompt and the question; the model (assistant role) then generates an initial step that ends with an EOS token. 
Each turn follows a \emph{reasoning–action–observation} loop:
(1) \textbf{Chain-of-thought (reasoning).} The Orchestrator analyzes the current state and plans the next action.
(2) \textbf{Tool call (action).} Based on its reasoning, Orchestrator selects a tool from the available set (e.g., APIs, specialized models, code interpreters) and specifies parameters. 
(3) \textbf{Tool response (observation).} If a tool call is present, the tool-call block is extracted and executed by the environment; the resulting output is appended to the context under the user role and fed back to the model for the next turn.
This process repeats until Orchestrator receives a termination signal from the environment or the rollout reaches a maximum of 50 turns.

\section{{\MethodName}}
\label{sec:main_method}

Our approach, {\MethodName}, centers on training a small language model as an intelligent agentic model capable of solving complex tasks by dynamically selecting and utilizing a wide variety of external tools. 
We hypothesize that small language models suffice for this purpose if they are taught to coordinate more intelligent tools strategically, and thus choose to train an 8B model.
{\MethodName} consists of an end-to-end reinforcement learning setup where the model under training, termed Orchestrator, learns to generate optimal multi-step reasoning and tool-use trajectories. 
The overall architecture is illustrated in Figure~\ref{fig:fig1}.

\vspace{-0.3em}
\subsection{Unified Tool Calling}
\label{sec:unified_tool_call}
\vspace{-0.3em}

In contrast to prior tool-use agents~\citep{li2025torl,jin2025search}, we broaden the toolset to include domain-specialized models and expose all tools through a single, unified interface.
Tools are specified in JSON as a list of objects; each object defines the tool name, description, and a typed parameter schema (names and descriptions). 
When LLMs are used as tools, we obtain their descriptions with the following steps: (1). randomly sample 10 training tasks; (2). obtain the trajectories of LLMs to finish these tasks; (3). Ask another LLM to write the description based on the task instructions, LLM trajectories and whether LLMs complete the tasks.
In Appendix~\ref{sec:app:model_description}, we show an example description of Qwen3-32B.
The complete catalog of tools used in our training is provided in Appendix~\ref{sec:app:tool_train}.

\subsection{End-to-End Agentic Reinforcement Learning}
\label{sec:end2endRL}
\paragraph{Reward design.} 
We introduce outcome, efficiency and preference rewards into the training.
For outcome reward, each rollout trajectory $\tau$ in a rollout batch $\Tau$ receives a binary accuracy reward $r_{\text{outcome}}(\tau) \in \{0,1\}$ 
based on whether $\tau$ solves the task: 
\begin{equation}
\small
r_{\text{outcome}}(\tau) = 
\begin{cases}
1 & \text{if } \text{solved}(\tau), \\
0 & \text{otherwise}.
\end{cases}
\end{equation}

We leverage GPT-5 as a judge to compare the answers, e.g., a name, a date, etc., providing greater flexibility in handling diverse predictions.

To encourage efficient solutions, we penalize the model under training for excessive compute or latency with the following rewards: $r_\text{compute}(\tau) = -\$(\tau)$, $r_\text{latency}(\tau) = -\mathit{Clock}(\tau)$, where $\$(\tau)$
is the monetary cost of $\tau$ and $\mathit{Clock}(\tau)$ is the consumed wall-clock time by $\tau$.
To establish a unified measurement on the compute of both open-sourced and proprietary models, we convert both the input tokens and output tokens to monetary costs following the third-party API pricing systems.
See more details in Appendix~\ref{sec:app:api_pricing}.

Preference reward is designed to encourage models to consider user preferences when choosing tools at each step.
Given a set of tools $\left\{t_1, t_2, ..., t_n\right\}$ and a rollout trajectory $\tau$, we consider the vector $M^{\tau}=[m^{\tau}_{t_1}, m^{\tau}_{t_2}, \ldots, m^{\tau}_{t_n},r_\text{outcome}(\tau),r_\text{compute}(\tau),r_\text{latency}(\tau)]$, where $m^{\tau}_{t_\bullet}$ is the number of times tool $t_\bullet$ is invoked in $\tau$, $M^{\tau}[n+1]=r_\text{outcome}(\tau)$.

During RL training, we normalize each element $M^{\tau}[k]$ for $1 \leq k \leq n+3$ over the rollout batch $\Tau$ as follows:
$M^{\tau}_\text{normalized}[k] = (M^{\tau}[k]-M^{\Tau}_{\text{min}}[k])/(M^\Tau_{\text{max}}[k]-M^\Tau_{\text{min}}[k])$, where $M^\Tau_{\text{min}}[k]$ and $M^\Tau_{\text{max}}[k]$
are minimum and maximum value for $M^{\bullet}[k]$ in the batch $\Tau$. 
If $M^\Tau_{\text{max}}[k]=M^\Tau_{\text{min}}[k]$, 
we disregard $M^{\tau}[k]$ by setting it to zero.
We calculate the final reward for a trajectory $\tau$ as:
\begin{equation}
\small
R(\tau) = 
\begin{cases}
M_{\text{normalized}}^{\tau} \cdot P & \text{if } r_{\text{outcome}}(\tau) \\
0 & \text{otherwise}.
\end{cases}
\label{eq:final_reward}
\end{equation}

where $P=\left[p_{t_1},p_{t_2}, ..., p_{t_n}, p_\text{outcome}, p_\text{compute}, p_\text{latency}\right]$ ($0 \leq p_{\bullet} \leq 1$) is the preference vector, indicating the extent the user would like to optimize $M[{\bullet}]$.
For example, $P[1] = p_{t_1}=1$ indicates strong user preference to use the tool $t_1$, while $P[n+1] = p_\text{outcome} = 1$ and $P[n+2] = p_\text{compute}=0$ implies that the user wants to exclusively optimize accuracy without considering the computational cost.

\paragraph{Training procedure.}
Orchestrator is fine-tuned using a policy gradient reinforcement learning algorithm, specifically Group Relative Policy Optimization (GRPO)~\citep{shao2024deepseekmath}. For each task in a batch, the policy $\pi_\theta$ generates a batch of trajectories ${\Tau}$. Each trajectory $\tau \in \Tau$ is assigned a scalar reward $R(\tau)$ (as calculated in Equation~\ref{eq:final_reward}), and GRPO normalizes this reward within its group to compute an advantage:
\begin{equation}
\small
A(\tau) = \frac{R(\tau) - \text{mean}_{\tau \in \Tau}{R(\tau)}}{\text{std}_{\tau \in \Tau}{R(\tau)}}.
\end{equation}
The policy is then updated to maximize the clipped surrogate objective:
\begin{equation}
\small
\mathcal{L}_{\text{GRPO}}(\theta) = \mathbb{E}{\tau \sim \pi_\theta} \Bigg[
\min\Big(
\text{ratio}_\theta(\tau) A(\tau), 
\text{clip}(\text{ratio}_\theta(\tau), 1 - \epsilon, 1 + \epsilon) A(\tau)
\Big)
\Bigg],
\end{equation}
where $\text{ratio}_\theta(\tau) = \frac{\pi_\theta(\tau)}{\pi_{\text{old}}(\tau)}$ is the likelihood ratio between the current and previous policy. 

\paragraph{Training techniques.}
To stabilize RL training and avoid KL loss explosion for this agent system, we perform the following during backward propagation: 
(1) \textit{homogeneity filtering}, when the standard deviation of rewards in a rollout batch is smaller than $0.1$, because this indicates that most rollouts in a batch exhibit similar behaviors, and provides weak training signals; 
(2) \textit{format consistency filtering}, when the example output is not aligned with the tool call format; 
(3) \textit{invalid output filtering}, when the example does not produce a valid answer or output.

\begin{figure}[t]
    \centering
    \includegraphics[width=0.95\linewidth]{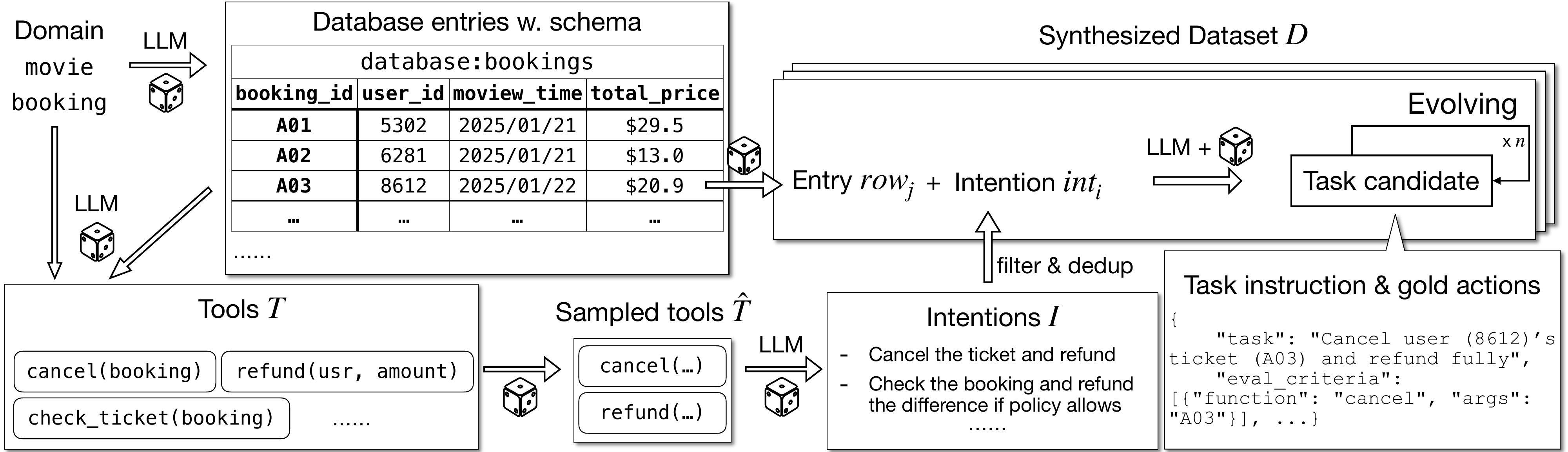}
    \vspace{-5pt}
    \caption{Overview of {\DataName} data synthesis pipeline. Starting from a domain, LLM will (1) firstly generate domain-specific database and tool APIs to simulate the environment and (2) then generate diverse user tasks together with their corresponding golden actions.}
    \label{fig:data_synthesis}
    \vspace{-10pt}
\end{figure}

\subsection{Data Synthesis}
\label{subsec:data_synthesis}
\paragraph{{\DataName}.}
To enable end-to-end RL training of Orchestrator, we require agentic tool-call tasks,
but verifiable data of this kind is scarce.
To generate such data, we devise a two-step process: 
(1) simulating rich user-agent-tool environments, including creating database schemas and tool APIs; and 
(2) generating diverse user tasks together with their corresponding ground truth solutions based on the environment.
Figure~\ref{fig:data_synthesis} provided an overview of this process.
Firstly, to simulate real-world user-agent-tool environments scalably, we choose a domain $D$ and then ask an LLM to generate a database which includes schema, major subjects to focus on and database entries (as illustrated in the top-left of Figure~\ref{fig:data_synthesis}).
Based on the given domain $D$, LLM proposes frequently-used tools.
Secondly, to increase the diversity of the task instructions, LLM first proposes diverse intents frequently seen in domain $D$, and then convert them to specific tasks based on detailed database information.
Each generated task consists of task instruction $I$, golden function calls $A={a_1, a_2, ..., a_l}$, and short information $o$ that must be mentioned during the process to solve the task.
To enhance the difficulty of the generated tasks, we leverage an additional LLM to complicate tasks by adding more complexities such as more constraints.
To ensure the quality of the synthesized data, we filter the data to remove a task if: 
(1) the execution of golden
function calls reports an error; 
(2) LLMs cannot solve it in pass@$8$; and 
(3) the task can be solved without any actions.
In Table~\ref{tab:data_statistics}, we list the statistics of the generated data in each domain.
More examples and prompts used to synthesize data can be found in Appendix~\ref{sec:app:data_synthesis}.
To evaluate whether a trajectory $\tau$ 
solves the given task, we define the following criteria: 
(1) \textit{execution correctness}, namely whether the database content matches after executing the golden function calls $A$ and the trajectory $\tau$;
(2) \textit{process fidelity}, i.e., whether the predefined information $o$, which is required to be communicated in the process to solve the task, is mentioned in $\tau$; 
(3) \textit{operation completeness}, that is whether the database entries operated in the ground truth trajectory $A$ are also operated in $\tau$.
We consider $\tau$ to solve the task if each of these three criteria is satisfied, and fail to solve it otherwise.

\paragraph{User preference.}
Different users possess different preferences.
For example, 
some users prefer local search to safeguard privacy, while others opt for internet-based search to access broader knowledge.
To train Orchestrator to account for such preferences in tool selection, we construct pairs of preference instruction $PI$ and preference vectors $P$, which indicate the extent a user would like to optimize certain features, e.g., latency, or the frequency to use a particular tool (\S\ref{subsec:data_synthesis}).
Given a tool set $\left\{t_1, t_2, ..., t_n\right\}$, and the corresponding configuration metadata (e.g., tool price, latency), an LLM proposes diverse pairs of $\left(PI,P\right)$, which are then valiadated by another LLM to verify consistency (see Appendix~\ref{sec:app:preference_example} for a sample pair).
The pairs are then split into two sets $\textit{Pairs}_{train}$ and $\textit{Pairs}_{eval}$ for training and evaluation, respectively.
We concatenate the generated preference instruction to the example instruction, and augment training and testing data with user preference. 
During training, we use Equation~\ref{eq:final_reward} and the generated preference vector $P$ to calculate reward, but using Equation~\ref{eq:final_reward_eval} and $P$ to calculate metrics in the evaluation. 
More details on rewards are included in  Appendix~\ref{sec:app:preference_reward_test}.

\paragraph{General tool configuration.}
To enhance Orchestrator’s generalization abilities, we curate a diverse set of tool configurations to prevent overfitting to specific usage patterns and encourage robust, general-purpose invocation. 
To emulate heterogeneous user access, we randomize the subset of tools available in each training instance, encouraging Orchestrator to optimize under varying constraints rather than relying on a fixed toolkit. 
We also vary pricing schedules across training instances to reflect heterogeneous tool costs, exposing the model to different cost configurations so it learns to adapt its optimization strategy as prices change. In aggregate, this approach models the variability in both tool availability and cost structures across users, yielding a richer supervisory signal for optimizing Orchestrator.

\section{Experimental Setting}
\label{sec:exp_settings}

\subsection{Tools}

\label{sec:tools}
In the training, we prepare a large and comprehensive tool set (Appendix~\ref{sec:app:tool_train}), but only sample a subset for each training instance to build diverse tool configurations (\S\ref{subsec:data_synthesis}).
We fix the following tool set in the evaluation for fair comparison.
\begin{itemize}
[leftmargin=*,label=$\bullet$,noitemsep,partopsep=0pt,topsep=0pt,parsep=0pt]
    \item \textbf{Basic tools.} We use Tavily search API~\footnote{\url{https://www.tavily.com/}} for web search, Python sandbox for Code interpreter, and build Faiss index with Qwen3-Embedding-8B~\citep{zhang2025qwen3} for local search. 
    Additionally, we also incorporate domain-specific functions, such as \texttt{get\_flight\_status}, to address specialized challenges within those domains.
    \item \textbf{Specialized LLMs.} We prompt GPT-5~\citep{gpt-5}, GPT-5-mini~\citep{gpt-5} as code writer, employ Qwen2.5-Coder-32B-Instruct~\citep{hui2024qwen2} as another code writer, and leverage Qwen2.5-Math-72B~\citep{yang2024qwen2}, Qwen2.5-Math-7B~\citep{yang2024qwen2} as specialized math models.
    \item \textbf{Generalist LLMs.} We consider GPT-5, GPT-5-mini, Llama-3.3-70B-Instruct~\citep{dubey2024llama}, and Qwen3-32B~\citep{yang2025qwen3} as representative generalist models.
\end{itemize}

\subsection{Baselines}

We compare Orchestrator-8B produced by {\MethodName} to baseline orchestrators constructed by prompting LLMs.
Additionally, we also compare to off-the-shelf monolithic LLM systems that are (1) not equipped with tools, (2) equipped with basic tools, and (3) using the expanded tool set that further includes specialized expert models and strong generalist models.

For off-the-shelf LLMs, we evaluate GPT-5, Claude Opus 4.1~\citep{claude41}, Llama-3.3-70B-Instruct, Qwen3-235B-A22B~\citep{yang2025qwen3}, Llama-3\_3-Nemotron-Super-49B-v1~\citep{bercovich2025llama}, Qwen3-8B~\citep{yang2025qwen3}.

\subsection{Evaluation Configuration}
We conduct experiments on three popular benchmarks with complex reasoning: \textbf{Humanity's Last Exam (HLE)}, \textbf{FRAMES}, and \textbf{$\tau^2$-Bench}. Details about these three benchmarks are given in Appendix~\ref{sec:app:evaluation_benchmarks}.
Throughout the evaluation, we use the official price for proprietary models and leverage the pricing systems of TogetherAI\footnote{\url{https://www.together.ai/pricing}} for open-source models. 
We set the inference temperature to 0 and allow maximum 50 turn for Orchestrator to solve a task.

\subsection{Training Configuration}

\label{sec:training_config}
We employ Qwen3-8B as the backbone LLM and train it on the GeneralThought-430K~\footnote{\url{https://huggingface.co/datasets/natolambert/GeneralThought-430K-filtered}} dataset in conjunction with synthetic data ($\S$\ref{subsec:data_synthesis}). 
The training configuration uses a learning rate of 1e-6, a maximum input sequence length of 24,000, and a maximum generation length of 8,000, with a training batch size of 16 and a rollout batch size of 8. 
We allow maximum 50 turns for the Orchestrator to finish a task during rollout and use 16 NVIDIA H100 GPUs throughout the training.

\begin{table*}[t]
\centering
\footnotesize
\caption{Comparison of Orchestrator-8B with baselines (prompt-based LLMs). 
Llama-Nemotron-49B denotes Llama-3.3-Nemotron-Super-49B-v1. 
Cost in US cents, Latency in minutes, are averaged between HLE and Frames. More efficiency statistics on $\tau^2$-Bench are in Table~\ref{tab:tau2_efficiency} in Appendix.
Basic tools include domain functions, search and code interpreter (\S\ref{sec:tools}).
$\uparrow$ The higher the better.
$\downarrow$ The lower the better. 
The results of existing SOTA are reported by ~\cite{gpt-5}$^\dagger$.
}

\label{tab:baseline_comparison}
\setlength{\tabcolsep}{4pt}
\begin{tabular}{@{} c p{2.9cm} c c c c c @{}}
\toprule
\textbf{Tools} & \textbf{Model(s)} & \textbf{HLE ($\uparrow$)} & \textbf{FRAMES ($\uparrow$)} & \textbf{$\tau^2$-Bench ($\uparrow$)} & \textbf{Cost ($\downarrow$)} & \textbf{Latency ($\downarrow$)} \\
\midrule
\multirow{3}{6em}{\centering \textcolor{gray}{Existing reported SOTA}}
  & \textcolor{gray}{GPT-5}   & \textcolor{gray}{35.2} & \textcolor{gray}{--} & \textcolor{gray}{84.2}$^\ddagger$ & \textcolor{gray}{--} & \textcolor{gray}{--} \\
  & \textcolor{gray}{o3}      & \textcolor{gray}{24.3} & \textcolor{gray}{--} & \textcolor{gray}{68.4} & \textcolor{gray}{--} & \textcolor{gray}{--} \\
  & \textcolor{gray}{GPT-4o}  & \textcolor{gray}{5.3}  & \textcolor{gray}{--} & \textcolor{gray}{43.8} & \textcolor{gray}{--} & \textcolor{gray}{--} \\
\midrule
\multirow{6}{6em}{\centering No tool}
  & Qwen3-8B              & 3.2  & 24.2  & --$^*$  & 0.2 & 0.6    \\
  & Llama-Nemotron-49B          & 3.6  & 25.6  & --$^*$ & 0.4 & 1.1   \\
  & Llama-3.3-70B         & 3.8  & 32.4 &  --$^*$ & 0.5 & 1.4  \\
  & Qwen3-235B-A22B       & 5.2  & 34.3 & --$^*$ & 2.6 & 3.3   \\
  & Claude Opus 4.1       & 11.7 & 58.2 & --$^*$ & 27.4 & 8.2  \\
  & GPT-5                 & 23.4 & 66.3 & --$^*$ & 6.2 & 4.1 \\
  
\midrule
\multirow{6}{7em}{\centering Basic tools}
  & Qwen3-8B                & 4.7   & 26.5  & 40.7  & 1.3 & 2.2 \\
  & Llama-Nemotron-49B            & 6.8   & 28.2  & 23.2  & 2.5 & 3.5 \\
  & Llama-3.3-70B           & 4.6   & 42.3  & 17.6  & 2.8 & 4.3  \\
  & Qwen3-235B-A22B         & 14.0  & 39.5  & 52.9  & 12.3  & 10.2 \\
  & Claude Opus 4.1         & 19.8  & 63.5  & 46.0  & 76.2 & 32.5  \\
  & GPT-5                   & 35.1  & 74.0  & 77.7  & 30.2 & 19.8 \\
\midrule
\multirow{7}{8em}{\centering Basic tools, \\ Specialized LLMs \\ Generalist LLMs}
  & Qwen3-8B                & 30.6  & 68.9  & 72.3  & 27.6 & 18.3  \\
  & Llama-Nemotron-49B            & 25.8  & 57.9  & 66.7  & 25.6 & 17.1  \\
  & Llama-3.3-70B           & 19.7  & 52.4  & 55.8  & 19.7 & 13.4  \\
  & Qwen3-235B-A22B         & 32.8  & 74.2  & 75.6  & 29.7 & 21.2 \\
  & Claude Opus 4.1         & 34.6  & 72.8  & 76.8  & 52.5 & 25.6 \\
  & GPT-5                   & 21.2  & 57.5  & 62.3  & 17.8 & 13.6 \\
  &\textbf{Orchestrator-8B} & \textbf{37.1}  & \textbf{76.3}  & \textbf{80.2}  & \textbf{9.2} & \textbf{8.2}   \\
\bottomrule
\end{tabular}

{\scriptsize
$^\dagger$ The HLE results of Existing reported SOTA are based on the full set, while other baselines and ours are only on the text-only subset. \\
$\ddagger$ Due to implementation differences, we could not fully reproduce GPT-5's reported result (84.2) and only reached 77.7 in our experiments. \\
$^*$ $\tau^2$-Bench cannot be solved in the absence of tools. 
}
\vspace{-0.2in}
\end{table*}

\vspace{-0.3em}
\section{Experimental Results}
\vspace{-0.3em}

We compare Orchestrator against a wide range of baselines across HLE, FRAMES, and $\tau^2$-Bench.
The results are summarized in Table~\ref{tab:baseline_comparison}.
For simple prompting methods without tools, models such as Qwen3-235B-A22B and Llama-3.3-70B fail to demonstrate strong performance.
This highlights the inherent difficulty of the benchmarks, where tool use or additional reasoning mechanisms is essential.
Providing tool access improves performance in some cases. 
For instance, Claude Opus 4.1 with tool usage improves from 11.7 to 19.8 in HLE, and from 58.2 to 63.5 in FRAMES, but at the expense of 2.8x costs and 4x latency.
Smaller models like Qwen3-8B perform poorly (4.7 on HLE), 
indicating that basic tools alone are insufficient. 
Combining tools with specialized and generalist LLMs generally improves results — Qwen3-235B-A22B, for example, rises from 14.0 to 32.8 on HLE and from 39.5 to 74.2 on FRAMES, but consumes more than 2 times of cost and latency.
However, the gains are inconsistent across different models. 
GPT-5 using both tools and models suffers from performance drop due to 
inherent biases, often defaulting to GPT-5-mini (\S\ref{sec:tool_use_analysis}).

In contrast, our Orchestrator-8B achieves 37.1 on HLE and 76.3 on FRAMES, surpassing all baselines by a large margin. 
In $\tau^2$-Bench, Orchestrator-8B outperforms GPT-5 using basic tools by 2.5\%, exhibiting strong function calling capabilities.
Notably, compared to GPT-5 with tool use (35.1 on HLE) and Qwen3-235B-A22B with tool + model (32.8 on HLE), our approach delivers consistent improvements of +2 to +4.3 absolute points, while using only a small fraction of cost and time. 
These gains are particularly striking given that Orchestrator has only 8B parameters, but is capable of coordinating more intelligent models such as GPT-5, and achieves better performance with lower cost, which highlights the effectiveness of the orchestration strategy.
Overall, the results clearly demonstrate the effectiveness of ToolOrchestra and the superiority of Orchestrator model in both performance and efficiency.

\section{Analysis}

\subsection{Tool Use Analysis}
\label{sec:tool_use_analysis}
\begin{figure}[t]
    \centering
    \includegraphics[width=\linewidth]{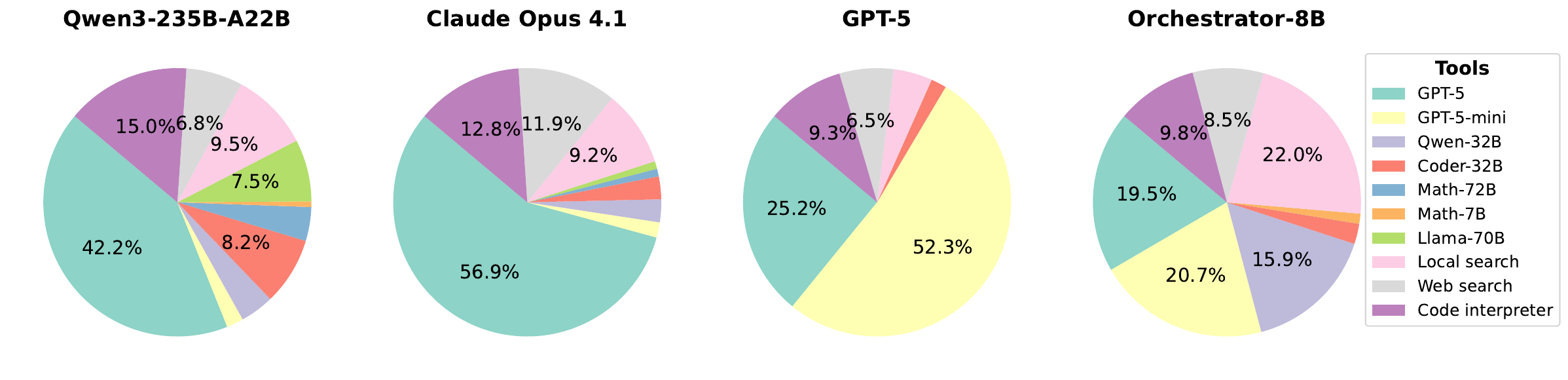}
    \vspace{-20pt}
    \caption{The proportion of tool calls made by LLMs to solve a task (averaged across HLE, Frames and $\tau^2$-bench).
Qwen-32B refers to Qwen3-32B~\citep{yang2025qwen3} and Coder-32B refers to Qwen2.5-Coder-32B-Instruct~\citep{hui2024qwen2}.
Compared to other strong foundation models, Orchestrator-8B makes more balanced tool calls, and does not exhibit strong biases toward a particular tool or model.
Detailed statistics are shown in Table~\ref{tab:tool_analysis_full}.
}
    \label{fig:tool_pies}
    \vspace{-0.2in}
\end{figure}

Figure~\ref{fig:tool_pies} shows the proportion of calls to each tool when various models serve as the orchestrator to solve a task.
Instead of excessively invoking strong models and expensive tools, Orchestrator-8B learns to coordinate them more strategically.
For example, in choosing between different models, Claude Opus 4.1 relies on GPT-5 most of the time, while making fewer calls to other models.
In contrast, GPT-5 prefers to use GPT-5-mini.
Orchestrator-8B learns to choose between various tools strategically, and achieves superior performance while using significantly lower costs.

\subsection{Cost Analysis}
\label{sec:cost_analysis}
\begin{wrapfigure}[14]{r}{0.4\columnwidth} 
  \vspace{-0.3in}
  \centering
  \includegraphics[width=0.35\columnwidth]{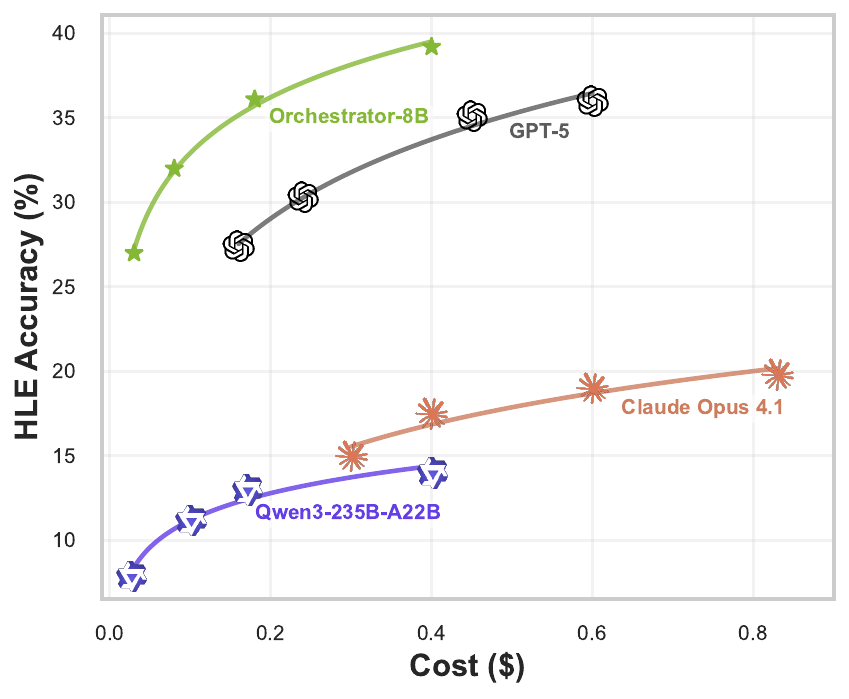}
  \vspace{-0.2in}
  \caption{The relationship between performance and cost. Compared to strong monolithic LLM systems, Orchestrator (ours) achieves the best cost-effectiveness.
  }
  \vspace{-0.1in}
  \label{fig:cost_effectiveness}
\end{wrapfigure}
To analyze the cost-effectiveness, we display the performance on HLE as a function of cost in Figure~\ref{fig:cost_effectiveness}. 
We experiment with settings where the maximum number of 10, 20, 50 and 100 turns are allowed to finish a task.
As the maximum number of allowed turns increases (i.e., cost increases), all models show improved performance. 
Orchestrator-8B consistently outperforms GPT-5, Claude Opus 4.1 and Qwen3-235B-A22B at a given budget, and can achieve similar results at a substantially lower cost.
This demonstrates the capability of Orchestrator-8B to manage a heterogeneous set of tools, and pushes the intelligence boundary of the system as a whole.

\subsection{Generalization}
\label{sec:main:generalization}
To evaluate Orchestrator-8B's generalization capability, we test it with a tool configuration containing models unseen during training:
(1) Query writer: Claude Opus 4.1, o3-mini and GPT-4o~\citep{gpt-4o};
(2) Code writer: Claude Opus 4.1, Claude Sonnet 4.1 and Codestral-22B-v0.1~\citep{codestral};
(3) Math model: OpenMath-Llama-2-70b~\citep{toshniwal2024openmath}, DeepSeek-Math-7b-Instruct~\citep{shao2024deepseekmath};
(4) Generalist Models:  Claude Opus 4.1, Claude Sonnet 4.1 and Gemma-3-27b-it~\citep{team2025gemma}.

\begin{wraptable}{r}{0.5\textwidth} 
\vspace{-0.1in}
\centering
\scriptsize
\caption{Generalization performance of Orchestrator-8B on HLE, Frames and $\tau^2$-Bench. 
}
\label{tab:model_generalization}
\resizebox{0.5\textwidth}{!}{%
\begin{tabular}{@{} p{2.7cm} c c c c c @{}}
\toprule
\textbf{Model(s)} & \textbf{HLE ($\uparrow$)} & \textbf{Frames ($\uparrow$)} & \textbf{$\tau^2$-Bench ($\uparrow$)} & \textbf{Cost ($\downarrow$)} & \textbf{Latency ($\downarrow$)} \\
\midrule
Qwen3-8B         & 12.6  & 34.9  & 38.3  & 37.9 & 10.6  \\
Llama-Nemotron-49B     & 13.9  & 32.7  & 22.9  & 53.6 & 8.3   \\
Llama-3.3-70B    & 13.2  & 49.3  & 12.8  & 63.3 & 10.1  \\
Qwen3-235B-A22B  & 14.7  & 63.5  & 38.7  & 87.2 & 13.8  \\
Claude Opus 4.1  & 17.8  & 53.6  & 43.4  & 102.4 & 19.5 \\
GPT-5            & 16.4  & 54.8  & 44.8  & 81.3 & 14.6 \\
\midrule
Orchestrator-8B  & \textbf{22.0}  & \textbf{73.8}  & \textbf{48.8}  & \textbf{34.8} & \textbf{8.2}   \\
\bottomrule
\end{tabular}
}
\end{wraptable}

We keep the basic tools (web search, local search and code interpreter) as the same mentioned in~\S\ref{sec:tools} and generate model descriptions following the same procedures mentioned in section \S\ref{sec:unified_tool_call}.
Table~\ref{tab:model_generalization} demonstrates that Orchestrator-8B shows strong skills in using models as tools.
Even provided with a set of models not seen during training, Orchestrator successfully adapts to it by understanding their skills and weaknesses from model descriptions, and consistently achieves the best performance at the lowest cost across HLE, Frames and $\tau^2$-Bench. 
This confirms that the diverse tool configurations during training effectively enhance the generalization capabilities of Orchestrator-8B.
In Appendix~\ref{sec:app:generalization_pricing}, we conduct further experiments to evaluate Orchestrator-8B on pricing configurations unseen in training.

\subsection{User Preferences}
\begin{wraptable}[12]{r}{0.45\textwidth}
\vspace{-0.4in}
\centering
\caption{Preference performance comparison. 
The results show that Orchestrator-8B best adapts to user preference during test time.}
\label{tab:preference}
\scriptsize
\resizebox{0.45\textwidth}{!}{%
\begin{tabular}{@{} p{2.7cm} c c c @{}}
\toprule
\textbf{Model(s)} & \textbf{HLE} & \textbf{Frames} & \textbf{$\tau^2$-Bench} \\
\midrule
Qwen3-8B         & 25.3  & 43.2  & 55.7 \\
Llama-Nemotron-49B     & 27.6  & 50.1  & 56.9 \\
Llama-3.3-70B    & 22.3  & 44.5  & 55.4 \\
Qwen3-235B-A22B  & 37.9  & 54.5  & 68.2 \\
Claude Opus 4.1  & 40.2  & 63.4  & 73.5 \\
GPT-5            & 34.6  & 62.3  & 70.3 \\
\midrule
Orchestrator-8B  & \textbf{46.7}  & \textbf{68.4}  & \textbf{79.5} \\
\bottomrule
\end{tabular}
}
\vspace{-0.2in}
\end{wraptable}
To assess Orchestrator-8B’s ability to adapt to heterogeneous user preferences at test time, we evaluate it on the Preference-aware test set described in \S\ref{subsec:data_synthesis}, where we concatenate each question with an additional preference instruction.
We evaluate the model preference adherence performance by calculating the preference-aware rewards defined in Appendix~\ref{sec:app:preference_reward_test}.
Table~\ref{tab:preference} shows that, even strong monolithic systems such as GPT-5 struggle to faithfully follow user preferences. 
In contrast, Orchestrator-8B exhibits remarkably better adherence to user preferences.

\section{Related Work}
\label{sec:related_work}
\subsection{From Tool Learning to Generalist Agents}
Tool learning underpins advanced reasoning in LLMs, as many tasks require external APIs, search engines, or code interpreters. 
Early work~\citep{schick2023toolformer, qintoolllm, qian2024tell} used supervised fine-tuning (SFT) on tool-labeled data like GPT-4 generated dialogues.
More recently, tool use has been framed as a sequential decision-making problem optimized by RL, with systems such as WebGPT~\citep{nakano2021webgpt}, Search-R1~\citep{jin2025search}, ToRL~\citep{li2025torl}, StepTool~\citep{yu2024steptool}, SWiRL~\citep{goldie2025synthetic}, Nemotron-Research-Tool-N1~\citep{zhang2025nemotron}, and ToolRL~\citep{qian2025toolrl}.
Building on this foundation, generalist agents like deep research agents~\citep{openai_deep_research_2025, deepmind_gemini_deep_research_2025, perplexity_deep_research_2025, moonshot_kimi_researcher_2025} autonomously discover, analyze, and synthesize across sources to produce analyst-level reports, aligning with the vision of compound AI systems~\citep{compound-ai-blog, chaudhry2025towards}. 
Recent open-source frameworks like SmolAgent~\citep{smolagents}, WebAgent~\citep{li2025websailor, wu2025webdancer, tao2025webshaper}, OWL~\citep{hu2025owl}, AutoAgent~\citep{tang2025autoagent}, and OAgent~\citep{zhu2025oagents} extend this trend toward modular, robust, and accessible systems, highlighting the broader democratization of generalist agents.

\subsection{From Tool-Use Accuracy to Efficiency and Controllability}
Beyond correctness, recent work emphasizes efficiency and controllability, aiming to reduce computational costs and better align outputs with user preferences. Prompting-based methods like Self Divide-and-Conquer~\citep{wang2025self}, Efficient Agents~\citep{wang2025efficient}, and SMART~\citep{qian2025smart} adaptively invoke tools or fine-tune usage, though they often depend on heavy prompt engineering or curated datasets. RL provides a more flexible alternative, where reward shaping balances accuracy, efficiency, and reliability. Advances include integrating auxiliary signals (e.g., response length, task difficulty)\citep{aggarwal2025l1, arora2025training, wang2025harnessing} and combining verifiable signals with human feedback\citep{peng2025agentic}.
A related direction is weak-to-strong generalization~\citep{burns2024weak}, which explores eliciting stronger models from weaker supervision. The most relevant work, OTC~\citep{wang2025otc}, improves efficiency by penalizing excessive calls while preserving accuracy. 
Unlike the prior work, our approach addresses the broader orchestration problem by using RL to coordinate diverse tools and models, balancing correctness, efficiency, and user preference for finer alignment and more robust deployment.

\section{Conclusion}
In this work, we presented {\MethodName}, a method for training a small orchestration model to unify diverse tools and specialized models. 
By training Orchestrator end-to-end with reinforcement learning, we showed that it can learn to plan adaptive tool-use strategies guided by both outcome quality, efficiency, and human preference rewards. 
This enables the agent to dynamically balance performance and cost, rather than relying on static heuristics or purely supervised approaches.
To aid reinforcement learning, we also contribute a complex user-agent-tool synthetic dataset {\DataName}.
Our experiments on challenging benchmarks
demonstrate that our Orchestrator-8B attains state-of-the-art performance while operating at significantly lower cost compared to larger models. 
Looking ahead, we envision more sophisticated recursive orchestrator systems to push the upper bound of intelligence but also to further enhance efficiency in solving increasingly complex agentic tasks.

\newpage
{
  \small
  \bibliographystyle{unsrt}
  \bibliography{main_tech_report}
}


\input{appendix}

\end{document}

%% file: math_commands.tex
\def\eqref#1{equation~\ref{#1}}

\def\1{\bm{1}}

\DeclareMathAlphabet{\mathsfit}{\encodingdefault}{\sfdefault}{m}{sl}
\SetMathAlphabet{\mathsfit}{bold}{\encodingdefault}{\sfdefault}{bx}{n}



%% file: appendix.tex
\newpage
\appendix


\section{Pilot Study}
\label{sec:app:pilot_study}

To evaluate the effectiveness of simple prompting as a method to configure an off-the-shelf language model to act as an orchestrator, we prompted GPT-5 and Qwen3-8B with a similar setting and the same prompt template we used in Section~\ref{sec:exp_settings}, allowing them to use GPT-5, GPT-5-mini, Qwen3-32B, and Qwen2.5-Coder-32B as tools and instruct the orchestrator to achieve best results while maintaining lowest cost. 
We then ran the model on a set of 300 HLE problems with max\_tokens=32,000  and temperature T=0 and monitored the average number of times each model referred to one of its model choices. 
The results are shown in Figure~\ref{fig:imbalanced-tool-calls}.
When Qwen3-8B serves as the orchestrator, it exhibits a strong tendency to delegate the task to GPT-5 (73\% of the cases).
We refer to this phenomenon as self-enhancement bias~\citep{zheng2023judging}, where the orchestrator favors its variants.
In contrast, when GPT-5 serves as the orchestrator, it prefers to call GPT-5 or GPT-5-mini in 98\% of the cases.
We term this phenomenon other-enhancement bias, where the orchestrator favors stronger models regardless of cost considerations, even though humans instruct them to do so.

Such imbalanced invocation patterns highlight the limitations of using off-the-shelf language models as orchestrators by simple prompting: their decisions are heavily biased rather than balanced across available options, resulting in poor orchestration effectiveness.
This observation motivates our method {\MethodName} to train a dedicated small orchestrator to decide when and how to invoke more intelligent language models.

\section{Evaluation Benchmarks}
\label{sec:app:evaluation_benchmarks}
\begin{itemize}
[leftmargin=*,label=$\bullet$,noitemsep,partopsep=0pt,topsep=0pt,parsep=0pt]
    \item \textbf{Humanity's Last Exam (HLE)}~\citep{phan2025humanity}. A large-scale benchmark comprising PhD-level questions across mathematics, humanities, natural sciences and more. 
    It evaluates the model capabilities to perform iterative search and intensive reasoning.
    Questions are multiple-choice or short-answer, with 10–14\% requiring images.  
    We use the text-only subset, designed to be unambiguous and not solvable by simple web search.
    \item \textbf{FRAMES}~\citep{krishna2024factfetchreasonunified}. 
    A dataset for end-to-end evaluation of retrieval-augmented generation (RAG), covering factuality, retrieval accuracy, and reasoning.  
    It contains 824 multi-hop questions requiring 2–15 Wikipedia articles, spanning numerical, tabular, temporal, and multi-constraint reasoning.
    \item \textbf{$\tau^2$-Bench}~\citep{barres2025tau}. 
    A benchmark to evaluate model capabilities to use tools and solve problems in conversations with users. It includes tasks in three domains: telecom, retail and airline.
\end{itemize}

\section{Model description for Qwen3-32B}
\label{sec:app:model_description}
The model shows advanced mathematical and quantitative reasoning, often solving complex problems and only faltering on highly specialized or computationally heavy items. 
Scientific domain knowledge is strong—especially in biology—with solid performance in physics and engineering; chemistry is mixed, with notable weaknesses in exact nomenclature and InChI outputs. 
Logical problem-solving is high, while humanities knowledge is moderate and uneven, with gaps in niche scholarly details. 
Coding and function call abilities are moderate, where it makes mistakes in parameters from time to time. 
Overall, the model has broad knowledge and robust analytic skills, but accuracy drops on narrow, recent, or rote-precision tasks, particularly in chemical informatics.

\section{Tools in training}
\label{sec:app:tool_train}
Below is the complete list of tools used in the training. For each example rollout, we randomly sample a subset of them to simulate heterogeneous availability of tools:
\begin{itemize}
    \item Query writer: GPT-5~\citep{gpt-5}, GPT-5-mini~\citep{gpt-5}, meta-llama/Llama-3.3-70B-Instruct~\citep{dubey2024llama}, meta-llama/Llama-3.1-8B-Instruct~\citep{dubey2024llama}, deepseek-ai/DeepSeek-R1~\citep{guo2025deepseek}, nvidia/Llama-3\_1-Nemotron-Ultra-253B-v1~\cite{bercovich2025llama}, microsoft/Phi-4-mini-instruct~\citep{abouelenin2025phi}, google/gemma-3-27b-it~\citep{team2025gemma}, Qwen/Qwen3-32B~\citep{yang2025qwen3}
    \item Web search: We use Tavily search API~\footnote{\url{https://www.tavily.com/}}  to provide orchestrator real-time web access.
    \item Local search: Faiss index with Qwen/Qwen3-Embedding-8B~\citep{zhang2025qwen3}
    \item Code writer + interpreter: We use GPT-5~\citep{gpt-5}, GPT-5-mini~\citep{gpt-5}, bigcode/starcoder2-15b~\citep{lozhkov2024starcoder}, and Qwen/Qwen2.5-Coder-32B-Instruct~\citep{hui2024qwen2} as code expert models to write code. We also implemented a Python sandbox to execute the code.
    \item Math models: Qwen/Qwen2.5-Math-72B~\citep{yang2024qwen2}, Qwen/Qwen2.5-Math-7B~\citep{yang2024qwen2}
    \item Generalist models: GPT-5~\citep{gpt-5}, GPT-5-mini~\citep{gpt-5}, meta-llama/Llama-3.3-70B-Instruct~\citep{dubey2024llama}, meta-llama/Llama-3.1-8B-Instruct~\citep{dubey2024llama}, deepseek-ai/DeepSeek-R1~\citep{guo2025deepseek}, nvidia/Llama-3\_1-Nemotron-Ultra-253B-v1~\cite{bercovich2025llama}, microsoft/Phi-4-mini-instruct~\citep{abouelenin2025phi}, Qwen/Qwen3-32B~\citep{yang2025qwen3}
\end{itemize}

\section{Third-party API}
\label{sec:app:api_pricing}
Here is a list of third-party APIs. We used pricing configurations for training:
\begin{itemize}
    \item TogetherAI: https://www.together.ai/
    \item Venice AI: https://docs.venice.ai/overview/about-venice
    \item Chutes: https://chutes.ai/
    \item NEBIUS: https://nebius.com/
    \item Lambda: https://lambda.ai/
    \item Hyperbolic: https://docs.hyperbolic.xyz/docs/welcome-to-hyperbolic
    \item Cloudflare: https://developers.cloudflare.com/
    \item Novita: https://novita.ai/
    \item AIML: https://aimlapi.com/
    \item Fireworks AI: https://fireworks.ai/
\end{itemize}
In the evaluation, we apply the pricing from Together AI for fair comparison.

\section{Humane preference example}
\label{sec:app:preference_example}
\textbf{Tools}; $T$ = $[$ Web search, local search, Qwen/Qwen3-235B-A22B, meta-llama/Llama-3.3-70B-Instruct, o3-mini, o3 $]$ \\
Preference instruction: $PI$ = I am a company employee and there is some confidential information in my server. There are many GPUs in the server, so I can host open-sourced models or retrievers. It would be great if we can avoid API calling as much as possible. \\
Preference vector: $P$ = [0,1,1,1,0,0,0,0,0]
Explanation: The first digit in the preference vector corresponds to the first tool in $T$; The second digit in the preference vector corresponds to the second tool in $T$, etc. The last three digits in $P$ corresponds to accuracy, cost and latency, aligned with the definitions in \S\ref{sec:end2endRL}. Therefore, this preference vector means the user prefers to use local search, Qwen/Qwen3-235B-A22B, meta-llama/Llama-3.3-70B-Instruct.

\section{Use of LLMs Disclosure}
We used GPT-5 to polish the writing, primarily in the Abstract, Introduction, Methodology, and Experiments sections, to improve the grammar, clarity, and readability.
The research ideas, methodology, experiments, and analyses were entirely conducted by the authors.

\section{Generalization of pricing configurations}
\label{sec:app:generalization_pricing}
In Section~\ref{sec:main:generalization}, we examined Orchestrator-8B’s ability to generalize to unseen tools. 
Here, we investigate its generalization across heterogeneous pricing regimes, where the same tools are assigned different costs. 
We evaluate whether the model adapts by adjusting its tool-calling strategy to optimize outcomes, efficiency, and alignment with user preferences—reflecting realistic settings in which tool costs vary across users. 
We test Orchestrator-8B under a pricing configuration not encountered during training. 
Specifically, we use the pricing configuration from deepinfra\footnote{https://deepinfra.com}.
As shown in Table~\ref{tab:generalization_pricing_config}, it consistently delivers superior performance, cost efficiency, and accuracy. These results demonstrate that training with diverse pricing configurations produces a model that is not constrained to a particular tool setup and can robustly generalize across diverse user scenarios.

\section{Data Synthesis}
\label{sec:app:data_synthesis}
\paragraph{{\DataName}.}
To enable end-to-end RL training of Orchestrator, we require data involving user-agent-tool interaction trajectories, but such verifiable data is scarce.
To generate such high-quality data, we devise a two-step process: 
(1) simulating rich user-agent-tool environments, including creating database schemas and tool APIs; and 
(2) based on the environment, generating diverse user tasks together with their corresponding ground truth solutions.
We further ensure quality by carefully verifying that each task is solvable using the provided databases and tool APIs.
Figure~\ref{fig:data_synthesis} provided an overview of this process.
Firstly, to simulate real-world user-agent-tool environments scalably, we choose a domain $D$ and then ask an LLM to generate a database which includes schema, major subjects to focus on and database entries (as illustrated in the top-left of Figure~\ref{fig:data_synthesis}).
Each entry is then checked to ensure coherence, adherence to the schema, and consistency across content, logic, and entities.
Based on the given domain $D$, LLM proposes frequently-used tools.
Secondly, to increase the diversity of the task instructions, LLM first proposes diverse intents frequently seen in domain $D$, which are later converted to specific tasks based on detailed database information.
Each generated task consists of task instruction $I$, gold function calls $A={a_1, a_2, ..., a_l}$, and short information $o$ that must be mentioned during the process to solve the task.
To enhance the difficulty of the generated tasks, we leverage an additional LLM to complicate tasks by adding more complexities such as more constraints.

To ensure the quality of the synthesized data, we filter the data to remove a task if: 
(1) the execution of golden
function calls reports an error; 
(2) LLMs cannot solve it in pass@$8$; and 
(3) the task can be solved without any actions.
In Appendix~\ref{sec:app:details_of_toolscale}, we list the statistics of the generated data in each domain.
More examples and prompts used to synthesize data can be found in Appendix~\ref{sec:app:data_synthesis_prompts}.
To evaluate whether a trajectory $\tau$ 
solves the given task, we define the following criteria: 
(1) \textit{execution correctness}, namely whether the database content matches after executing the golden function calls $A$ and the trajectory $\tau$;
(2) \textit{process fidelity}, i.e., whether the predefined information $o$, which is required to be communicated in the process to solve the task, is mentioned in $\tau$; 
(3) \textit{operation completeness}, that is whether the database entries operated in the ground truth trajectory $A$ are also operated in $\tau$.
We consider $\tau$ solves the task if all of three criteria are satisfied, or fails otherwise.

\begin{table*}[t]
\centering
\scriptsize
\caption{Generalization performance under different a pricing configuration. Orchestrator-8B consistently performs the best in terms of performance, cost and latency, which illustates the robustness of the Orchestrator}
\label{tab:generalization_pricing_config}
\setlength{\tabcolsep}{4pt}
\begin{tabular}{@{} c c c c c c @{}}
\toprule
& \textbf{HLE ($\uparrow$)} & \textbf{Frames ($\uparrow$)} & \textbf{$\tau^2$-Bench ($\uparrow$)} & \textbf{Cost ($\downarrow$)} & \textbf{Latency ($\downarrow$)} \\

\midrule
Qwen3-8B                & 29.7  & 68.2  & 71.9  & 27.4 & 17.9  \\
Nemotron-49B            & 25.6  & 57.8  & 66.3  & 24.3 & 17.2  \\
Llama-3.3-70B           & 19.6  & 52.2  & 55.4  & 17.9 & 12.0  \\
Qwen3-235B-A22B         & 32.4  & 74.1  & 75.3  & 27.9 & 20.8 \\
Claude Opus 4.1         & 34.5  & 72.3  & 76.4  & 52.3 & 25.1 \\
GPT-5                   & 20.8  & 57.3  & 61.9  & 17.5 & 13.2 \\
\textbf{Orchestrator-8B} & \textbf{36.9}  & \textbf{76.6}  & \textbf{80.4}  & \textbf{7.5} & \textbf{7.8}   \\
\bottomrule
\end{tabular}
\vspace{-0.2in}
\end{table*}

\section{Breakdown of {\DataName}}
\label{sec:app:details_of_toolscale}
\begin{table*}[h!]
\centering
\caption{Statistics of {\DataName}: the number of tools, database entries, and tasks per domain. 
}
\label{tab:data_statistics}
\setlength{\tabcolsep}{4pt}
\resizebox{\textwidth}{!}{
\begin{tabular}{@{} c c c c c c c c c c c @{}}
\toprule
 & Finanace & Sport & E-commerce & Medicine & Entertainment  & Railway   & Restaurant & Education  & Travel  & Weather \\
\midrule
Tools &  22 & 19 & 15 & 19 & 24 & 25 &  23 & 21 & 15 & 14  \\
DB Entries & 686 & 423 & 577 & 920 & 852 & 790 & 683 & 816 & 752 & 549  \\
Tasks &  396 & 247 & 343 & 622 & 561 & 414 & 348 & 426 & 331 & 375   \\
\bottomrule
\end{tabular}
}
\end{table*}

\section{Data synthesis prompts and examples}
\label{sec:app:data_synthesis_prompts}
\begin{table}[ht]
\caption{Model prompts to generate subjects in a domain}
\centering
\begin{tabular}{p{13cm}}
\toprule
Generate a list of major subjects in \{domain\}. \\
Output using the following format: \\
\texttt{\textasciigrave\textasciigrave\textasciigrave} \\
$[$subject1, subject2, ...$]$ \\
\texttt{\textasciigrave\textasciigrave\textasciigrave} \\
\bottomrule
\end{tabular}
\label{tab:prompt_generate_subject}
\end{table}

\begin{table}[ht]
\caption{Model prompts to generate schema in a domain}
\centering
\begin{tabular}{p{13cm}}
\toprule
\texttt{\textasciigrave\textasciigrave\textasciigrave} \\
\{demo\_schema\} \\
\texttt{\textasciigrave\textasciigrave\textasciigrave} \\
Generate another schema of similar formats for \{domain\}. \\
\bottomrule
\end{tabular}
\label{tab:prompt_generate_schema}
\end{table}

\begin{table}[ht]
\caption{Model prompts to generate database entry}
\centering
\begin{tabular}{p{13cm}}
\toprule
Schema \\
\texttt{\textasciigrave\textasciigrave\textasciigrave} \\
\{schema\} \\
\texttt{\textasciigrave\textasciigrave\textasciigrave} \\
\bigbreak
Following the schema, write records in the subject \{subject\}. Make sure that the values align with the definitions in the schema and are consistent in different places. Use the following format to output: \\
\texttt{\textasciigrave\textasciigrave\textasciigrave} \\
\{ ``...": ..., ``...": ..., \} \\
\texttt{\textasciigrave\textasciigrave\textasciigrave} \\
Wrap the dictionary within \texttt{\textasciigrave\textasciigrave\textasciigrave}.\\
\bottomrule
\end{tabular}
\label{tab:prompt_generate_db_entry}
\end{table}

\begin{table}[ht]
\caption{Model prompts to validate database entry}
\centering
\begin{tabular}{p{13cm}}
\toprule
Schema \\
\texttt{\textasciigrave\textasciigrave\textasciigrave} \\
\{schema\} \\
\texttt{\textasciigrave\textasciigrave\textasciigrave} \\
\bigbreak
Database entry \\
\texttt{\textasciigrave\textasciigrave\textasciigrave} \\
\{db\_entry\} \\
\texttt{\textasciigrave\textasciigrave\textasciigrave} \\
\bigbreak
Please check whether the following conditions are satisfied: \\
Condition 1. The database entry strictly aligns with the fields and type definitions in the schema. \\
Condition 2. The values in the database entry are consistent across different places, e.g., id, name, etc. \\
Condition 3. The database content is logical, natural, and reasonable. \\
Output using the following format: \\
\texttt{\textasciigrave\textasciigrave\textasciigrave} \\
\{ ``condition 1": ``satisfied or not satisfied, ``condition 2": ``satisfied or not satisfied, ``condition 3": ``satisfied or not satisfied, \} \\
\texttt{\textasciigrave\textasciigrave\textasciigrave} \\
\bottomrule
\end{tabular}
\label{tab:prompt_validate_db_entry}
\end{table}

\begin{table}[ht]
\caption{Model prompts to generate functions}
\centering
\begin{tabular}{p{13cm}}
\toprule
Demonstration function \\
\texttt{\textasciigrave\textasciigrave\textasciigrave} \\
\{demo\_function\} \\
\texttt{\textasciigrave\textasciigrave\textasciigrave} \\
\bigbreak
Following the formats of demonstration function, write frequently-used functions in \{domain\}. Wrap the functions within \texttt{\textasciigrave\textasciigrave\textasciigrave}.\\
\bottomrule
\end{tabular}
\label{tab:prompt_generate_functions}
\end{table}

\begin{table}[ht]
\caption{Model prompts to generate intents}
\centering
\begin{tabular}{p{13cm}}
\toprule
Functions \\
\texttt{\textasciigrave\textasciigrave\textasciigrave} \\
\{functions\} \\
\texttt{\textasciigrave\textasciigrave\textasciigrave} \\
\bigbreak
Propose realistic intents in \{domain\} that could be solved by the functions above.
Use the following format to output: \\
\texttt{\textasciigrave\textasciigrave\textasciigrave}.\\
$[$ \\
    ``purpose 1", \\
    ``purpose 2", \\
    ... \\
$]$ \\
\texttt{\textasciigrave\textasciigrave\textasciigrave}.\\
\bottomrule
\end{tabular}
\label{tab:prompt_generate_intents}
\end{table}

\begin{table}[ht]
\caption{Model prompts to generate tasks}
\centering
\begin{tabular}{p{13cm}}
\toprule
Functions \\
\texttt{\textasciigrave\textasciigrave\textasciigrave} \\
\{functions\} \\
\texttt{\textasciigrave\textasciigrave\textasciigrave} \\
Database \\
\texttt{\textasciigrave\textasciigrave\textasciigrave} \\
\{database\} \\
\texttt{\textasciigrave\textasciigrave\textasciigrave} \\
\bigbreak
Propose a realistic task with the intent: \{intent\}. Use the following format to output: \\
\texttt{\textasciigrave\textasciigrave\textasciigrave}.\\
\{task\_template\} \\
\texttt{\textasciigrave\textasciigrave\textasciigrave}.\\
\bottomrule
\end{tabular}
\label{tab:prompt_generate_tasks}
\end{table}

\begin{table}[ht]
\caption{Model prompts to evolve tasks}
\centering
\begin{tabular}{p{13cm}}
\toprule
Functions \\
\texttt{\textasciigrave\textasciigrave\textasciigrave} \\
\{functions\} \\
\texttt{\textasciigrave\textasciigrave\textasciigrave} \\
Database \\
\texttt{\textasciigrave\textasciigrave\textasciigrave} \\
\{database\} \\
\texttt{\textasciigrave\textasciigrave\textasciigrave} \\
Old task: \{task\} \\
\bigbreak
Make a new task by adding more complexity to the old task. You can add constraints, involve more entities, make the situation more tricky, etc. Use the following format to output: \\
\texttt{\textasciigrave\textasciigrave\textasciigrave}.\\
\{task\_template\} \\
\texttt{\textasciigrave\textasciigrave\textasciigrave}.\\
\bottomrule
\end{tabular}
\label{tab:prompt_evolve_tasks}
\end{table}

\begin{table}[ht]
\caption{Database schema example}
\centering
\begin{tabular}{p{13cm}}
\toprule
\{ \\
    ``movies": \{ \\
        ``MMMMMMM": \{    \/\/ movie\_id \\
            ``movie\_id": "MMMMMMM", \\
            ``title": "...", \\
            ``genres": $[$``Action", ``Adventure", ``Comedy", ``Drama", ``Horror", ``Thriller", ``Romance", ``Science Fiction", ``Fantasy", ``Mystery"$]$,\\
            ``runtime\_minutes": ...,\\
            ``mpaa\_rating": ``...",\\
            ``languages\_audio": $[$"..."$]$,\\
            ``subtitles": $[$"..."$]$,\\
            ``formats": $[$"2D", "3D", "IMAX", "Dolby"$]$,\\
            ``release\_date": ``YY-MM-DD",\\
            ``end\_of\_run\_est": ``YY-MM-DD",\\
            ``cast": $[$\\
                \{ ``name": ``...", ``role": ``..." \}\\
            $]$,\\
            ``crew": \{ \\
                ``director": ``...", \\
                ``writer": ``...", \\
                ``producer": ``...", \\
                ``music": ``..." \\
            \}, \\
            "synopsis": "..." \\
        \}, \\
        ... \\
    \}, \\
    ... \\
\} \\
\bottomrule
\end{tabular}
\label{tab:schema_example}
\end{table}

\clearpage
\section{Calculation of rewards for preference-aware benchmark}
\label{sec:app:preference_reward_test}
During training, we directly follow the Equation~\ref{eq:final_reward} to calculate rewards.
In the evaluation, we use the following procedure.
Following the definition in \S\ref{sec:end2endRL}, we have a tool set $\left\{t_1, t_2, ..., t_n\right\}$ and a rollout trajectory $\tau$, we consider the vector $M^{\tau}=[m^{\tau}_{t_1}, m^{\tau}_{t_2}, \ldots, m^{\tau}_{t_n},r^{\tau}_\text{outcome},r^{\tau}_\text{compute},r^{\tau}_\text{latency}]$, where $m^{\tau}_{t_\bullet}$ is the number of times tool $t_\bullet$ is invoked in $\tau$.
In the evaluation, we obtain the baseline vector $M_0$ by running the starting checkpoint, e.g., Qwen3-8B.
For example, if we would like to evaluate a checkpoint $\mathit{CKPT}_s$ that is trained for $s$ steps from a base Qwen3-8B model, then we first run Qwen3-8B on the benchmark and record the vector $M^{\tau(e)}_0$ as the baseline vector for the Qwen3-8B's trajectory $\tau(e)$ for each example $e$ of the benchmark.
We then obtain $M^{\tau(e)}_s$ by running $\mathit{CKPT}_s$ on the same example $e$.
$M^{\tau(e)}_s$ is normalized as 
\begin{equation}
M^{{\tau(e)}}_{\text{normalized}, s}[k] = 
\begin{cases}
M^{\tau(e)}_s[k]/max(1,M^{\tau(e)}_0[k]) & \text{if } 1 \leq k \leq n+1 \\
M^{\tau(e)}_0[k]/max(1,M^{\tau(e)}_s[k]) & \text{otherwise}.
\end{cases}
\label{eq:normalize_reward_eval}
\end{equation}
We then proceed to calculate the final preference-aware reward for the example $e$ as:
\begin{equation}
\small
R^e(\tau) = 
\begin{cases}
M^{{\tau(e)}}_{\text{normalized}, s}\cdot P & \text{if } r_{\text{outcome}(\tau)} \\
0 & \text{otherwise}.
\end{cases}
\label{eq:final_reward_eval}
\end{equation}

The benchmark result is calculated as the sum of $R^e(\tau)$ over the examples $e$ of the benchmark.

\begin{table*}[t]
\centering
\caption{
The average number of calls on each tool when various models serve as the orchestrator to solve an instance (averaged across HLE, Frames and $\tau^2$-bench).
Qwen-32B refers to Qwen/Qwen3-32B~\citep{yang2025qwen3}, Coder-32B refers to Qwen/Qwen2.5-Coder-32B-Instruct~\citep{hui2024qwen2}, Math-7B refers to https://huggingface.co/Qwen/Qwen2.5-Math-7B-Instruct~\citep{yang2024qwen2}, Math-72B refers Qwen/Qwen2.5-Math-72B-Instruct~\citep{yang2024qwen2}, and Llama-70B refers to meta-llama/Llama-3.3-70B-Instruct~\citep{dubey2024llama}.
Compared to other strong foundation models, Orchestrator-8B achieves better results (Table~\ref{tab:baseline_comparison}) while making few calls to GPT-5.
}
\label{tab:tool_analysis_full}
\small
\setlength{\tabcolsep}{4pt}
\resizebox{\textwidth}{!}{
\begin{tabular}{@{} c c c c c c c c c c c c c c  @{}}
\toprule
Model & GPT-5 & GPT-5-mini & Qwen-32B & Coder-32B & Math-72B & Math-7B & Llama-70B & Local search & Web search & Code interpreter \\
\midrule
Qwen3-8B & 6.0 & 0.5 & 1.4 & 0.5 & 0.0 & 0.0 & 0.0 & 0.8 & 1.2 & 1.6 \\
Nemontron-49B & 5.1 & 1.6 & 0.5 & 0.8 & 0.1 & 0.1 & 0.3 & 0.7 & 0.9 & 1.4 \\
Llama-3.3-70B   & 1.8 & 0.0 & 0.1 & 0.0 & 1.4 & 0.3 & 4.8 & 0.6 & 1.4 & 1.3 \\
Qwen3-235B-A22B & 6.2 & 0.3 & 0.6 & 1.2 & 0.6 & 0.1 & 1.1 & 1.4 & 1.0 & 2.2 \\
Claude Opus 4.1 & 6.2 & 0.2 & 0.3 & 0.3 & 0.1 & 0.0 & 0.1 & 1.0 & 1.3 & 1.4   \\
GPT-5           & 2.7 & 5.6 & 0.0 & 0.2 & 0.0 & 0.0 & 0.0 & 0.5 & 0.7 & 1.0 \\
Orchestrator-8B & 1.6 & 1.7 & 1.3 & 0.2 & 0.0 & 0.1 & 0.0 & 1.8 & 0.7 & 0.8\\

\bottomrule
\end{tabular}
}
\end{table*}

\begin{table*}[t]
\centering
\scriptsize
\caption{The cost and latency of LLMs in $\tau^2$-Bench. Orchestrator-8B consistently shows better performance with lower cost and latency.}
\label{tab:tau2_efficiency}
\setlength{\tabcolsep}{4pt}
\begin{tabular}{@{} c p{2.7cm} c c c @{}}
\toprule
\textbf{Tools} & \textbf{Model(s)} & \textbf{$\tau^2$-Bench ($\uparrow$)} & \textbf{Cost ($\downarrow$)} & \textbf{Latency ($\downarrow$)} \\
  
\midrule
\multirow{6}{7em}{\centering Basic tools}
  & Qwen3-8B                & 40.7  & 1.6 & 2.3 \\
  & Llama-Nemotron-49B      & 23.2  & 2.7 & 3.6 \\
  & Llama-3.3-70B           & 17.6  & 3.1 & 4.5  \\
  & Qwen3-235B-A22B         & 52.9  & 12.6  & 10.6 \\
  & Claude Opus 4.1         & 46.0  & 81.2 & 32.8  \\
  & GPT-5                   & 77.7  & 31.3 & 20.2 \\
\midrule
\multirow{7}{8em}{\centering Basic tools, \\ Specialized LLMs \\ Generalist LLMs}
  & Qwen3-8B                & 72.3  & 27.9 & 18.4  \\
  & Llama-Nemotron-49B      & 66.7  & 25.8 & 17.5  \\
  & Llama-3.3-70B           & 55.8  & 20.1 & 14.2  \\
  & Qwen3-235B-A22B         & 75.6  & 30.0 & 22.6 \\
  & Claude Opus 4.1         & 76.8  & 52.8 & 24.1 \\
  & GPT-5                   & 62.3  & 18.2 & 14.5 \\
  &\textbf{Orchestrator-8B} & \textbf{80.2}  & \textbf{10.3} & \textbf{8.6}   \\
\bottomrule
\end{tabular}
\vspace{-0.2in}
\end{table*}

%% file: main_tech_report.bbl
\begin{thebibliography}{10}

\bibitem{phan2025humanity}
Long Phan, Alice Gatti, Ziwen Han, Nathaniel Li, Josephina Hu, Hugh Zhang, Chen Bo~Calvin Zhang, Mohamed Shaaban, John Ling, Sean Shi, et~al.
\newblock Humanity's last exam.
\newblock {\em arXiv preprint arXiv:2501.14249}, 2025.

\bibitem{qintoolllm}
Yujia Qin, Shihao Liang, Yining Ye, Kunlun Zhu, Lan Yan, Yaxi Lu, Yankai Lin, Xin Cong, Xiangru Tang, Bill Qian, et~al.
\newblock Toolllm: Facilitating large language models to master 16000+ real-world apis.
\newblock In {\em The Twelfth International Conference on Learning Representations}, 2023.

\bibitem{schick2023toolformer}
Timo Schick, Jane Dwivedi-Yu, Roberto Dess{\`\i}, Roberta Raileanu, Maria Lomeli, Eric Hambro, Luke Zettlemoyer, Nicola Cancedda, and Thomas Scialom.
\newblock Toolformer: Language models can teach themselves to use tools.
\newblock {\em Advances in Neural Information Processing Systems}, 36:68539--68551, 2023.

\bibitem{qin2024tool}
Yujia Qin, Shengding Hu, Yankai Lin, Weize Chen, Ning Ding, Ganqu Cui, Zheni Zeng, Xuanhe Zhou, Yufei Huang, Chaojun Xiao, et~al.
\newblock Tool learning with foundation models.
\newblock {\em ACM Computing Surveys}, 57(4):1--40, 2024.

\bibitem{gehring2024rlef}
Jonas Gehring, Kunhao Zheng, Jade Copet, Vegard Mella, Quentin Carbonneaux, Taco Cohen, and Gabriel Synnaeve.
\newblock Rlef: Grounding code llms in execution feedback with reinforcement learning.
\newblock {\em arXiv preprint arXiv:2410.02089}, 2024.

\bibitem{qian2024tell}
Cheng Qian, Bingxiang He, Zhong Zhuang, Jia Deng, Yujia Qin, Xin Cong, Zhong Zhang, Jie Zhou, Yankai Lin, Zhiyuan Liu, et~al.
\newblock Tell me more! towards implicit user intention understanding of language model driven agents.
\newblock In {\em Proceedings of the 62nd Annual Meeting of the Association for Computational Linguistics (Volume 1: Long Papers)}, pages 1088--1113, 2024.

\bibitem{yu2024steptool}
Yuanqing Yu, Zhefan Wang, Weizhi Ma, Shuai Wang, Chuhan Wu, Zhiqiang Guo, and Min Zhang.
\newblock Steptool: Enhancing multi-step tool usage in llms through step-grained reinforcement learning.
\newblock {\em arXiv preprint arXiv:2410.07745}, 2024.

\bibitem{goldie2025synthetic}
Anna Goldie, Azalia Mirhoseini, Hao Zhou, Irene Cai, and Christopher~D Manning.
\newblock Synthetic data generation \& multi-step rl for reasoning \& tool use.
\newblock {\em arXiv preprint arXiv:2504.04736}, 2025.

\bibitem{zhang2025nemotron}
Shaokun Zhang, Yi~Dong, Jieyu Zhang, Jan Kautz, Bryan Catanzaro, Andrew Tao, Qingyun Wu, Zhiding Yu, and Guilin Liu.
\newblock Nemotron-research-tool-n1: Exploring tool-using language models with reinforced reasoning.
\newblock {\em arXiv preprint arXiv:2505.00024}, 2025.

\bibitem{qian2025toolrl}
Cheng Qian, Emre~Can Acikgoz, Qi~He, Hongru Wang, Xiusi Chen, Dilek Hakkani-T{\"u}r, Gokhan Tur, and Heng Ji.
\newblock Toolrl: Reward is all tool learning needs.
\newblock {\em arXiv preprint arXiv:2504.13958}, 2025.

\bibitem{zheng2023judging}
Lianmin Zheng, Wei-Lin Chiang, Ying Sheng, Siyuan Zhuang, Zhanghao Wu, Yonghao Zhuang, Zi~Lin, Zhuohan Li, Dacheng Li, Eric Xing, et~al.
\newblock Judging llm-as-a-judge with mt-bench and chatbot arena.
\newblock {\em Advances in neural information processing systems}, 36:46595--46623, 2023.

\bibitem{barres2025tau}
Victor Barres, Honghua Dong, Soham Ray, Xujie Si, and Karthik Narasimhan.
\newblock {${\tau}^2$-Bench: Evaluating Conversational Agents in a Dual-Control Environment}.
\newblock {\em arXiv preprint arXiv:2506.07982}, 2025.

\bibitem{krishna2024factfetchreasonunified}
Satyapriya Krishna, Kalpesh Krishna, Anhad Mohananey, Steven Schwarcz, Adam Stambler, Shyam Upadhyay, and Manaal Faruqui.
\newblock Fact, fetch, and reason: A unified evaluation of retrieval-augmented generation, 2024.

\bibitem{belcak2025small}
Peter Belcak, Greg Heinrich, Shizhe Diao, Yonggan Fu, Xin Dong, Saurav Muralidharan, Yingyan~Celine Lin, and Pavlo Molchanov.
\newblock Small language models are the future of agentic ai.
\newblock {\em arXiv preprint arXiv:2506.02153}, 2025.

\bibitem{zhao2025llm}
Bingxi Zhao, Lin~Geng Foo, Ping Hu, Christian Theobalt, Hossein Rahmani, and Jun Liu.
\newblock Llm-based agentic reasoning frameworks: A survey from methods to scenarios.
\newblock {\em arXiv preprint arXiv:2508.17692}, 2025.

\bibitem{xi2024agentgym}
Zhiheng Xi, Yiwen Ding, Wenxiang Chen, Boyang Hong, Honglin Guo, Junzhe Wang, Dingwen Yang, Chenyang Liao, Xin Guo, Wei He, et~al.
\newblock Agentgym: Evolving large language model-based agents across diverse environments.
\newblock {\em arXiv preprint arXiv:2406.04151}, 2024.

\bibitem{zhou2024archer}
Yifei Zhou, Andrea Zanette, Jiayi Pan, Sergey Levine, and Aviral Kumar.
\newblock Archer: Training language model agents via hierarchical multi-turn rl.
\newblock {\em arXiv preprint arXiv:2402.19446}, 2024.

\bibitem{xi2025agentgym}
Zhiheng Xi, Jixuan Huang, Chenyang Liao, Baodai Huang, Honglin Guo, Jiaqi Liu, Rui Zheng, Junjie Ye, Jiazheng Zhang, Wenxiang Chen, et~al.
\newblock Agentgym-rl: Training llm agents for long-horizon decision making through multi-turn reinforcement learning.
\newblock {\em arXiv preprint arXiv:2509.08755}, 2025.

\bibitem{li2025torl}
Xuefeng Li, Haoyang Zou, and Pengfei Liu.
\newblock Torl: Scaling tool-integrated rl.
\newblock {\em arXiv preprint arXiv:2503.23383}, 2025.

\bibitem{jin2025search}
Bowen Jin, Hansi Zeng, Zhenrui Yue, Jinsung Yoon, Sercan Arik, Dong Wang, Hamed Zamani, and Jiawei Han.
\newblock Search-r1: Training llms to reason and leverage search engines with reinforcement learning.
\newblock {\em arXiv preprint arXiv:2503.09516}, 2025.

\bibitem{shao2024deepseekmath}
Zhihong Shao, Peiyi Wang, Qihao Zhu, Runxin Xu, Junxiao Song, Xiao Bi, Haowei Zhang, Mingchuan Zhang, YK~Li, Yang Wu, et~al.
\newblock Deepseekmath: Pushing the limits of mathematical reasoning in open language models.
\newblock {\em arXiv preprint arXiv:2402.03300}, 2024.

\bibitem{zhang2025qwen3}
Yanzhao Zhang, Mingxin Li, Dingkun Long, Xin Zhang, Huan Lin, Baosong Yang, Pengjun Xie, An~Yang, Dayiheng Liu, Junyang Lin, et~al.
\newblock Qwen3 embedding: Advancing text embedding and reranking through foundation models.
\newblock {\em arXiv preprint arXiv:2506.05176}, 2025.

\bibitem{gpt-5}
OpenAI.
\newblock Introducing gpt-5.
\newblock \url{https://openai.com/index/introducing-gpt-5/}.
\newblock Accessed: 2025-09-23.

\bibitem{hui2024qwen2}
Binyuan Hui, Jian Yang, Zeyu Cui, Jiaxi Yang, Dayiheng Liu, Lei Zhang, Tianyu Liu, Jiajun Zhang, Bowen Yu, Keming Lu, et~al.
\newblock Qwen2. 5-coder technical report.
\newblock {\em arXiv preprint arXiv:2409.12186}, 2024.

\bibitem{yang2024qwen2}
An~Yang, Beichen Zhang, Binyuan Hui, Bofei Gao, Bowen Yu, Chengpeng Li, Dayiheng Liu, Jianhong Tu, Jingren Zhou, Junyang Lin, et~al.
\newblock Qwen2. 5-math technical report: Toward mathematical expert model via self-improvement.
\newblock {\em arXiv preprint arXiv:2409.12122}, 2024.

\bibitem{dubey2024llama}
Abhimanyu Dubey, Abhinav Jauhri, Abhinav Pandey, Abhishek Kadian, Ahmad Al-Dahle, Aiesha Letman, Akhil Mathur, Alan Schelten, Amy Yang, Angela Fan, et~al.
\newblock The llama 3 herd of models.
\newblock {\em arXiv e-prints}, pages arXiv--2407, 2024.

\bibitem{yang2025qwen3}
An~Yang, Anfeng Li, Baosong Yang, Beichen Zhang, Binyuan Hui, Bo~Zheng, Bowen Yu, Chang Gao, Chengen Huang, Chenxu Lv, et~al.
\newblock Qwen3 technical report.
\newblock {\em arXiv preprint arXiv:2505.09388}, 2025.

\bibitem{claude41}
{Anthropic}.
\newblock Claude opus 4.1.
\newblock \url{https://www.anthropic.com/news/claude-opus-4-1}, 2025.

\bibitem{bercovich2025llama}
Akhiad Bercovich, Itay Levy, Izik Golan, Mohammad Dabbah, Ran El-Yaniv, Omri Puny, Ido Galil, Zach Moshe, Tomer Ronen, Najeeb Nabwani, et~al.
\newblock Llama-nemotron: Efficient reasoning models.
\newblock {\em arXiv preprint arXiv:2505.00949}, 2025.

\bibitem{gpt-4o}
OpenAI.
\newblock Hello gpt-4o.
\newblock \url{https://openai.com/index/hello-gpt-4o/}.
\newblock Accessed: 2025-09-23.

\bibitem{codestral}
Mistral~AI team.
\newblock Codestral.
\newblock \url{https://mistral.ai/news/codestral}, 2024.

\bibitem{toshniwal2024openmath}
Shubham Toshniwal, Ivan Moshkov, Sean Narenthiran, Daria Gitman, Fei Jia, and Igor Gitman.
\newblock Openmathinstruct-1: A 1.8 million math instruction tuning dataset.
\newblock {\em arXiv preprint arXiv: Arxiv-2402.10176}, 2024.

\bibitem{team2025gemma}
Team {Google}, Aishwarya Kamath, Johan Ferret, Shreya Pathak, Nino Vieillard, Ramona Merhej, Sarah Perrin, Tatiana Matejovicova, Alexandre Ram{\'e}, Morgane Rivi{\`e}re, et~al.
\newblock Gemma 3 technical report.
\newblock {\em arXiv preprint arXiv:2503.19786}, 2025.

\bibitem{nakano2021webgpt}
Reiichiro Nakano, Jacob Hilton, Suchir Balaji, Jeff Wu, Long Ouyang, Christina Kim, Christopher Hesse, Shantanu Jain, Vineet Kosaraju, William Saunders, et~al.
\newblock Webgpt: Browser-assisted question-answering with human feedback.
\newblock {\em arXiv preprint arXiv:2112.09332}, 2021.

\bibitem{openai_deep_research_2025}
{OpenAI}.
\newblock Introducing deep research, 2025.

\bibitem{deepmind_gemini_deep_research_2025}
{Google DeepMind}.
\newblock Gemini deep research — your personal research assistant, 2025.

\bibitem{perplexity_deep_research_2025}
{Perplexity AI}.
\newblock Introducing perplexity deep research, 2025.

\bibitem{moonshot_kimi_researcher_2025}
{Moonshot AI}.
\newblock Kimi-researcher: End-to-end rl training for emerging agentic capabilities, 2025.

\bibitem{compound-ai-blog}
Matei Zaharia, Omar Khattab, Lingjiao Chen, Jared~Quincy Davis, Heather Miller, Chris Potts, James Zou, Michael Carbin, Jonathan Frankle, Naveen Rao, and Ali Ghodsi.
\newblock The shift from models to compound ai systems.
\newblock \url{https://bair.berkeley.edu/blog/2024/02/18/compound-ai-systems/}, 2024.

\bibitem{chaudhry2025towards}
Gohar~Irfan Chaudhry, Esha Choukse, {\'I}{\~n}igo Goiri, Rodrigo Fonseca, Adam Belay, and Ricardo Bianchini.
\newblock Towards resource-efficient compound ai systems.
\newblock In {\em Proceedings of the 2025 Workshop on Hot Topics in Operating Systems}, pages 218--224, 2025.

\bibitem{smolagents}
Aymeric Roucher, Albert~Villanova del Moral, Thomas Wolf, Leandro von Werra, and Erik Kaunismäki.
\newblock `smolagents`: a smol library to build great agentic systems.
\newblock \url{https://github.com/huggingface/smolagents}, 2025.

\bibitem{li2025websailor}
Kuan Li, Zhongwang Zhang, Huifeng Yin, Liwen Zhang, Litu Ou, Jialong Wu, Wenbiao Yin, Baixuan Li, Zhengwei Tao, Xinyu Wang, et~al.
\newblock Websailor: Navigating super-human reasoning for web agent.
\newblock {\em arXiv preprint arXiv:2507.02592}, 2025.

\bibitem{wu2025webdancer}
Jialong Wu, Baixuan Li, Runnan Fang, Wenbiao Yin, Liwen Zhang, Zhengwei Tao, Dingchu Zhang, Zekun Xi, Gang Fu, Yong Jiang, et~al.
\newblock Webdancer: Towards autonomous information seeking agency.
\newblock {\em arXiv preprint arXiv:2505.22648}, 2025.

\bibitem{tao2025webshaper}
Zhengwei Tao, Jialong Wu, Wenbiao Yin, Junkai Zhang, Baixuan Li, Haiyang Shen, Kuan Li, Liwen Zhang, Xinyu Wang, Yong Jiang, et~al.
\newblock Webshaper: Agentically data synthesizing via information-seeking formalization.
\newblock {\em arXiv preprint arXiv:2507.15061}, 2025.

\bibitem{hu2025owl}
Mengkang Hu, Yuhang Zhou, Wendong Fan, Yuzhou Nie, Bowei Xia, Tao Sun, Ziyu Ye, Zhaoxuan Jin, Yingru Li, Qiguang Chen, et~al.
\newblock Owl: Optimized workforce learning for general multi-agent assistance in real-world task automation.
\newblock {\em arXiv preprint arXiv:2505.23885}, 2025.

\bibitem{tang2025autoagent}
Jiabin Tang, Tianyu Fan, and Chao Huang.
\newblock Autoagent: A fully-automated and zero-code framework for llm agents.
\newblock {\em arXiv preprint arXiv:2502.05957}, 2025.

\bibitem{zhu2025oagents}
He~Zhu, Tianrui Qin, King Zhu, Heyuan Huang, Yeyi Guan, Jinxiang Xia, Yi~Yao, Hanhao Li, Ningning Wang, Pai Liu, et~al.
\newblock Oagents: An empirical study of building effective agents.
\newblock {\em arXiv preprint arXiv:2506.15741}, 2025.

\bibitem{wang2025self}
Hongru Wang, Boyang Xue, Baohang Zhou, Tianhua Zhang, Cunxiang Wang, Huimin Wang, Guanhua Chen, and Kam-Fai Wong.
\newblock Self-dc: When to reason and when to act? self divide-and-conquer for compositional unknown questions.
\newblock In {\em Proceedings of the 2025 Conference of the Nations of the Americas Chapter of the Association for Computational Linguistics: Human Language Technologies (Volume 1: Long Papers)}, pages 6510--6525, 2025.

\bibitem{wang2025efficient}
Ningning Wang, Xavier Hu, Pai Liu, He~Zhu, Yue Hou, Heyuan Huang, Shengyu Zhang, Jian Yang, Jiaheng Liu, Ge~Zhang, et~al.
\newblock Efficient agents: Building effective agents while reducing cost.
\newblock {\em arXiv preprint arXiv:2508.02694}, 2025.

\bibitem{qian2025smart}
Cheng Qian, Emre~Can Acikgoz, Hongru Wang, Xiusi Chen, Avirup Sil, Dilek Hakkani-T{\"u}r, Gokhan Tur, and Heng Ji.
\newblock Smart: Self-aware agent for tool overuse mitigation.
\newblock {\em arXiv preprint arXiv:2502.11435}, 2025.

\bibitem{aggarwal2025l1}
Pranjal Aggarwal and Sean Welleck.
\newblock L1: Controlling how long a reasoning model thinks with reinforcement learning.
\newblock {\em arXiv preprint arXiv:2503.04697}, 2025.

\bibitem{arora2025training}
Daman Arora and Andrea Zanette.
\newblock Training language models to reason efficiently.
\newblock {\em arXiv preprint arXiv:2502.04463}, 2025.

\bibitem{wang2025harnessing}
Rui Wang, Hongru Wang, Boyang Xue, Jianhui Pang, Shudong Liu, Yi~Chen, Jiahao Qiu, Derek~Fai Wong, Heng Ji, and Kam-Fai Wong.
\newblock Harnessing the reasoning economy: A survey of efficient reasoning for large language models.
\newblock {\em arXiv preprint arXiv:2503.24377}, 2025.

\bibitem{peng2025agentic}
Hao Peng, Yunjia Qi, Xiaozhi Wang, Zijun Yao, Bin Xu, Lei Hou, and Juanzi Li.
\newblock Agentic reward modeling: Integrating human preferences with verifiable correctness signals for reliable reward systems.
\newblock {\em arXiv preprint arXiv:2502.19328}, 2025.

\bibitem{burns2024weak}
Collin Burns, Pavel Izmailov, Jan~Hendrik Kirchner, Bowen Baker, Leo Gao, Leopold Aschenbrenner, Yining Chen, Adrien Ecoffet, Manas Joglekar, Jan Leike, et~al.
\newblock Weak-to-strong generalization: Eliciting strong capabilities with weak supervision.
\newblock In {\em International Conference on Machine Learning}, pages 4971--5012. PMLR, 2024.

\bibitem{wang2025otc}
Hongru Wang, Cheng Qian, Wanjun Zhong, Xiusi Chen, Jiahao Qiu, Shijue Huang, Bowen Jin, Mengdi Wang, Kam-Fai Wong, and Heng Ji.
\newblock Otc: Optimal tool calls via reinforcement learning.
\newblock {\em arXiv e-prints}, pages arXiv--2504, 2025.

\bibitem{guo2025deepseek}
Daya Guo, Dejian Yang, Haowei Zhang, Junxiao Song, Ruoyu Zhang, Runxin Xu, Qihao Zhu, Shirong Ma, Peiyi Wang, Xiao Bi, et~al.
\newblock Deepseek-r1: Incentivizing reasoning capability in llms via reinforcement learning.
\newblock {\em arXiv preprint arXiv:2501.12948}, 2025.

\bibitem{abouelenin2025phi}
Abdelrahman Abouelenin, Atabak Ashfaq, Adam Atkinson, Hany Awadalla, Nguyen Bach, Jianmin Bao, Alon Benhaim, Martin Cai, Vishrav Chaudhary, Congcong Chen, et~al.
\newblock Phi-4-mini technical report: Compact yet powerful multimodal language models via mixture-of-loras.
\newblock {\em arXiv preprint arXiv:2503.01743}, 2025.

\bibitem{lozhkov2024starcoder}
Anton Lozhkov, Raymond Li, Loubna~Ben Allal, Federico Cassano, Joel Lamy-Poirier, Nouamane Tazi, Ao~Tang, Dmytro Pykhtar, Jiawei Liu, Yuxiang Wei, et~al.
\newblock Starcoder 2 and the stack v2: The next generation.
\newblock {\em arXiv preprint arXiv:2402.19173}, 2024.

\end{thebibliography}
